\documentclass[12pt]{article}

\usepackage[linesnumbered,ruled,vlined]{algorithm2e}
\usepackage[utf8]{inputenc}
\usepackage{multirow}
\usepackage{array}
\usepackage{amssymb,bbm, cancel}
\usepackage{amsmath}
\usepackage{amsthm,comment}
\theoremstyle{definition}

\newtheorem{prop}{Proposition}

\DeclareMathOperator*{\argmax}{argmax}

\usepackage{caption}
\usepackage{subcaption,graphicx}
\usepackage[margin=1in]{geometry}
\usepackage{tabularx}
\usepackage{setspace, comment}
\usepackage{parskip}
\usepackage{graphicx}
\usepackage[round]{natbib}
\usepackage{hyperref}
\usepackage{mathabx}
\usepackage{graphicx}
\usepackage{siunitx}
\usepackage{dsfont}
\usepackage[table]{xcolor}
\newcommand{\our}{NA$_0$CT$^2$ }
\graphicspath{ {./images/} }
\usepackage{float} 
\usepackage{appendix}

\title{Noise-Augmented $\ell_0$ Regularization of Tensor Regression with Tucker Decomposition}
\author{Tian Yan,   Yinan Li, Fang Liu\footnote{corresponding author (fang.liu.131@nd.edu)} \\
\normalsize{Applied and Computational Mathematics and Statistics}\\
\normalsize{University of Notre Dame, Notre Dame, IN 46556, USA}}
\date{} 

\begin{document}
\maketitle\vspace{-1em}
\begin{abstract}
Tensor data are multi-dimension arrays. Low-rank decomposition-based regression methods with tensor predictors exploit the structural information in tensor predictors while significantly reducing the number of parameters in tensor regression.  We propose a  method named \our (Noise Augmentation for $\ell_0$ regularization on Core Tensor in Tucker decomposition) to regularize the parameters in  tensor regression (TR), coupled with Tucker decomposition. We establish theoretically that \our  achieves exact $\ell_0$ regularization  on the core tensor from the Tucker decomposition in linear TR and generalized linear TR. To our knowledge, \our is the first Tucker decomposition-based regularization method in TR to achieve $\ell_0$ in core tensors.  \our is implemented through an iterative procedure and involves two straightforward steps in each iteration -- generating noisy data based on the core tensor from the Tucker decomposition of the updated parameter estimate and running a regular GLM on noise-augmented data on vectorized predictors. We demonstrate the implementation of \our and its $\ell_0$ regularization effect in both simulation studies and real data applications. The results suggest that \our can improve predictions compared to other decomposition-based TR approaches, with or without regularization and it identifies important predictors though not designed for that purpose.\vspace{9pt}

\noindent \textbf{keywords:} tensor regression (TR), low-rank decomposition, CP decomposition, Tucker decomposition, vectorization, $\ell_0$ regularization, sparsity,  noise augmentation, generalized linear model
\end{abstract}

\section{Introduction}
\setstretch{1.05}
Tensor data are $D$-dimensional arrays in the $\mathbb{R}^{I_1\times I_2\times...I_D}$ space. Many applications, such as medical imaging and recommendation systems, contain tensor data.  Vectors are 1D tensors ($D=1$) and matrices are 2D tensors ($D=2$). A na\"{i}ve approach for analyzing tensor data  is to vectorize tensors and apply traditional regression models to the vectorized data. Not only do regression models on 
vectorized data involve a number of parameters in this approach, but the structural information contained in the original tensors also is not leveraged during the estimation. To overcome these limitations, tensor regression (TR) models are  proposed by applying low-rank decomposition to tensors, including the CANDECOMP/PARAFAC (CP) decomposition \citep{harshman1970foundations, carroll1970analysis} and the Tucker decomposition \citep{tucker1966some}, that would reduce the number of parameters in regression models  (e.g. from tens of millions down to a few hundred \citep{zhou2013tensor}) while retaining the spatial structure information in tensor data. Even with low-rank decomposition, the number of parameters ($p$) in TR may still be large relative to the sample size ($n$). When $n<p$, regularizations on TR parameters are necessary for the parameter to be identifiable; even if $n>p$, regularizations are often employed to improve the generalization of  estimated TR. 

\cite{guo2011tensor} apply CP decomposition in linear TR and regularize parameters with Frobenius norm and group-sparsity norm. \citet{zhou2013tensor} apply CP decomposition and suggest that commonly used regularizers for  generalized linear models (GLMs) can be also applied to TR, such as the b$\ell_2$, $\ell_2$, lasso,  elastic net, and SCAD regularizers. \cite{signoretto2014learning} extend  spectral regularization to TR with convex and differentiable loss functions. \cite{liu2014trace} impose Frobenius norm on the difference between a full tensor and its CP decomposition and  trace norm on its CP components to achieve relaxed rank regularization  and use alternating direction method of multipliers (ADMM) for optimization \citep{boyd2011distributed}.  
\cite{wimalawarne2016theoretical} investigate regularization based on overlapped trace norm and (scaled) latent trace norm and employ dual optimization to estimate parameters. \cite{song2017multilinear} impose both nuclear and $\ell_1$ norms in linear TR to achieve a mixture of rank and $\ell_1$ regularization. 
\cite{li2018tucker} employ Tucker decomposition and regularize the core tensor with common regularizers for GLMs. 
\cite{raskutti2019convex} assume that tensor parameters are located in a low-dimensional subspace and convex and weakly decomposable (a relaxed regularity condition on regularizers to study their oracle properties; see \citet{van2014weakly} regularizers, and  propose a convex optimization framework for multi-dimensional responses. 
The rank of a low-rank decomposition in TR is often assumed known or pre-specified. If the rank is unknown or not prespecified, the method by \cite{he2018boosted} can be applied that assumes sparsity in each unit-rank tensor in CP decomposition and conducts unit-rank tensor regression iteratively.  \cite{ou2020sparse} impose sparsity regularization on coefficient tensors directly rather than  factor matrices from the CP decomposition of the former, and use ADMM for optimization. \cite{roy2022regularized}  decompose a coefficient tensor into a low tubal rank tensor and a sparse tensor and solve a convex optimization problem using ADMM. \cite{chen2022effective} consider the correlation between different slices of a coefficient tensor, and propose latent F-1 norm that set an entire slice of the tensor at zero to promote sparsity.  \cite{xu2022graph} develop a graph regularizer to incorporate domain knowledge (intra-modal relations) in TR that is encoded in a graph Laplacian matrix and apply the approach to financial data.

Many of the methods mentioned above assume convexity on loss functions. \cite{chen2019non} propose a non-convex projected gradient descent algorithm with error bounds on estimated tensors and show  superior performance in both statistical error and run-time in examples. $\ell_0$ regularization is also non-convex.  To our knowledge, there is little work that explores the $\ell_0$ regularization in TR. $\ell_0$ regularization is a desirable regularizer in that it sets zero-valued parameters exactly at zero while providing unbiased estimates for non-zero parameters. On the other hand, $\ell_0$ regularization is NP-hard. In the setting of GLMs, approximate $\ell_0$ regularizations through continuous functions are often used, such as \citet{wei2018gradient, tang2014sparse, hyder2009approximate, LiGang, Lin, lv2009unified, shi2018admm}, among others. Some of these methods are examined only when the loss function is least squares and do not offer theoretical guarantees that the approximated $\ell_0$ regularizations achieve both accurate variable selection and unbiased estimates for non-zero parameters, and some of them would require the development of new optimization procedures to achieve more efficient  computation  or more accurate results. To our knowledge, these approximate $\ell_0$ regularizers have not been examined in TR with CP or Tucker decompositions.

In this study, we propose to achieve $\ell_0$ regularization in TR using the noise-augmented (NA) regularization technique.  NA has been used for regularizing the estimation and inferences in GLMs estimation and construction of single and multiple undirected graphs \citep{li2021adaptive, li2022adaptive}. It works by augmenting the original data with noisy samples drawn from Gaussian distributions with adaptive variance terms  designed to achieve different types of regularizations such as the $\ell_1$, $\ell_2$, elastic net, SCAD, group lasso, as well as $\ell_0$ regularizations among others. 

We refer to our method as \emph{NA$_0$CT$^2$ (Noise Augmentation for $\ell_0$ regularization on Core Tensor in Tucker decomposition)}. The core tensor in a Tucker decomposition serves as a central and compressed representation of the interactions among the latent factors extracted along the modes of the original tensor and is central to understanding the relationships between the factors along each mode. To our knowledge, NA$_0$CT$^2$ is the first to explore the $\ell_0$ regularization on the core tensor with the Tucker decomposition in TR. The achieved sparsity on the core tensor elements through NA$_0$CT$^2$ signifies the absence of interaction between specific latent factors and can drastically simplify the interpretation: help identify important variables that matter in the prediction  and interpret high-dimensional TR models.  In addition, NA$_0$CT$^2$ is easy to implement and involves a simple noisy data generation step followed by running regular GLMs (without regularizers) on noise-augmented data iteratively, where one can leverage existing software.

In what follows, we first introduce some basic concepts on tensor decomposition and TR in Section \ref{sec:prelim}. We present the NA$_0$CT$^2$  method and an algorithm to implement it in Section \ref{sec:method}. We apply NA$_0$CT$^2$ to both simulated tensor data and two real tensor datasets in Section \ref{sec:example}. The paper concludes in Section \ref{sec:discu} with some final remarks on NA$_0$CT$^2$ and provides future research directions.

\section{Preliminaries}\label{sec:prelim}

Let $x$ denote a scalar,  $\mathbf{x}$ denote a vector, $\mathbf{X}$ denote a matrix, and $\mathcal{X}$ denote a $I_1\times I_2\times\dots\times I_D$ tensor, where $D$ is the dimensionality  or the \emph{order} of the tensor, the $d$-th dimension is known as the \emph{$d$-mode} for $d=1,\ldots,D$, and $I_d$ is the dimension of the $d$-mode. For example, a typical Magnetic Resonance Imaging (MRI) image  is of size $256\times256\times256$. If expressed as a  tensor, it is of order $3$ and $I_d=256$ for $d=1,2,3$. We use
$x_{i,j}$ to denote the element $(i,j)$ in matrix $\mathbf{X}$, and $x_i$ to denote  element $i$ in vector $\mathbf{x}$, and $x_{i_1i_2\dots i_D}$ to denote the element in position $(i_1, i_2, \dots, i_D)$ in $\mathcal{X}\in\mathbb{R}^{I_1\times I_2\times\ldots\times I_D}$.

A tensor can be formulated as an \textit{outer product} of $D$ vectors as in $\mathcal{X}=\mathbf{a^{(1)}}\circ\mathbf{a^{(2)}}\circ\dots\circ\mathbf{a^{(D)}}$, where $\mathbf{a^{(i)}}\in\mathbb{R}^{I_i}$ for $1\leqslant i\leqslant D$; the element $x_{i_1i_2\dots i_D}$ in  $\mathcal{X}$ is given by $a^{(1)}_{i_1}a^{(2)}_{i_2}\dots a^{(D)}_{i_D}$ for $1\leqslant i_d\leqslant i_D$. The \textit{vectorization} of $\mathcal{X}\in \mathbb{R}^{I_1\times I_2\times\dots\times I_D}$ is denoted by $\mbox{vec}(\mathcal{X})$, and  $x_{i_1i_2\dots i_D}$ is the $(1+\sum_{d=1}^{D}(i_d-1)\prod_{d'=1}^{d-1}I_{d'})$-th element of $\mbox{vec}(\mathcal{X})$ (when $d=1, \prod_{d'=1}^{d-1}I_{d'}$ is replaced by 1).  %
The \textit{$d$-mode matricization} of $\mathcal{X}\in \mathbb{R}^{I_1\times I_2\times\dots\times I_D}$ is denoted as $\mathbf{X}_{(d)}\in\mathbb{R}^{I_d\times (I_1I_2\dots I_{d-1}I_{d+1}\dots I_D)}$ and  $x_{i_1i_2\dots i_D}$ is entry $(i_d,j)$ in matrix $\mathbf{X}_{(d)}$, where $j=1+\sum_{k=1,k\neq d}^{D}(i_k-1)\prod_{m=1,m\neq d}^{k-1}I_{m}$ (when $k=1, \prod_{m=1,m\neq d}^{k-1}I_{m}$ is replaced by 1). %
The \textit{$d$-mode product} of tensor $\mathcal{X}\in \mathbb{R}^{I_1\times I_2\times\dots\times I_D}$ and matrix $\mathbf{A}\in\mathbb{R}^{J\times I_d}$, denoted by $\mathcal{X}\times_d\mathbf{A}\in\mathbb{R}^{I_1\times I_2\times\dots\times I_{d-1}\times J\times I_{d+1}\times\dots\times I_D}$, is given by $(\mathcal{X}\times_d\mathbf{A})_{i_1i_2\dots i_{d-1}ji_{d+1}\dots i_D}=\sum_{i_d=1}^{I_d}x_{i_1i_2\dots i_D}a_{ji_d}$. The \textit{inner product} of two tensors $\mathcal{X}$ and $\mathcal{Y}$  of the same order $D$ is  $\langle \mathcal{X},\mathcal{Y}\rangle$ that is calculated as $\sum^{I_1}_{i_1=1}\sum^{I_2}_{i_2=1}\dots\sum^{I_D}_{i_D=1} x_{i_1i_2\dots i_D}y_{i_1i_2\dots i_D}$.

In TR,  predictors $\mathcal{X}_i\in\mathbb{R}^{I_1\times I_2\times \dots \times I_D}$ are tensors and observations $\mathbf{y}=(y_1,y_2,\dots,y_n)$ on response variable $Y$ are assumed to be independently identically distributed. If $Y$ belongs to exponential family distributions, then TR models can be formulated as follows,
\begin{equation}\label{eq:glm}
p(y_i|\mathcal{X}_i,\mathcal{B},\phi)=\mbox{exp}\bigg\{\frac{y_i\langle\mathcal{X}_i,\mathcal{B}\rangle-B(\langle\mathcal{X}_i,\mathcal{B}\rangle)}{a(\phi)}+h(y_i,\phi)\bigg\},
\end{equation}
where $\mathcal{B}\in\mathbb{R}^{I_1\times I_2\times \dots \times I_D}$ contains unknown parameters and is a tensor of the same order as $\mathcal{X}$. The most straightforward way to run the model in Eqn \eqref{eq:glm} with a tensor predictor to vectorize $\mathcal{X}$ and $\mathcal{B}$ and run GLM with or without regularization on the vectorized $\mathcal{B}$. However, the number of parameters $\prod_{i=d}^D I_d$ in  $\mathcal{B}$ is usually large, causing difficulty in both estimation and computation, as well as in the interpretability of the estimated model. In addition, the structural information in $\mathcal{X}$ is lost. To reduce the number of parameters while retaining some structural information in the tensor data, TR can be run with a low-rank decomposition of  $\mathcal{B}$, such as Tucker decomposition or the CP decomposition, instead of a fully-parameterized  $\mathcal{B}$.

A \emph{Tucker decomposition} of tensor $\mathcal{B}\in \mathbb{R}^{I_1\times I_2\times\dots\times I_D}$ is defined as
\begin{align}
\mathcal{B}&\approx\mathcal{G}\times_1\mathbf{U}_1\times_2\mathbf{U}_2\times_3\dots\times_\mathrm{D}\mathbf{U}_D=\ldbrack\mathcal{G};\mathbf{U}_1,\mathbf{U}_2,\dots,\mathbf{U}_D\rdbrack\notag\\
&=\textstyle\sum_{r_1=1}^{R_1}\sum_{r_2=1}^{R_2}\dots\sum_{r_D=1}^{R_D}g_{r_1r_2\dots r_D}\mathbf{u}_1^{r_1}\circ\mathbf{u}_2^{r_2}\circ\dots\circ\mathbf{u}_D^{r_D}, \label{eq:tucker}
\end{align}
where $\mathcal{G}\in \mathbb{R}^{R_1\times R_2\times\dots\times R_D}$ is the \emph{core tensor} with $R_d\leq I_d$, $\mathbf{U}_d\in\mathbb{R}^{I_d\times R_d}$ is a matrix with orthonormal columns for $d=1,\ldots, D$, and $\mathbf{u}_d^{r_d}$ is the $r_d$-th column of matrix $\mathbf{U}_d$ for $d=1,\ldots, D$. When $R_d= I_d$, the Tucker decomposition  in Eqn \eqref{eq:tucker} is an exact representation of the original tensor and 
``$\approx$'' can be replaced by ``$=$''. When $R_d< I_d$, the core tensor is often viewed as a compressed representation of the original tensor.
The Tucker decomposition can be regarded as a form of higher-order principal component analysis; $\mathbf{U}_d$ is the ``principle component'' (latent factor) extracted along the $d$-mode and the elements  in the core tensor measure presents of the interactions among the latent factors.

An rank-$R$ \emph{CP decomposition} of $\mathcal{B}$ is \vspace{-12pt}
\begin{equation}\label{eq:CP}
\mathcal{B}\approx\ldbrack\mathbf{g};\mathbf{U}_1,\mathbf{U}_2,\dots,\mathbf{U}_D\rdbrack
=\textstyle\sum_{r=1}^{R}g_r\mathbf{u}_1^{r}\circ\mathbf{u}_2^{r}\circ\dots\circ\mathbf{u}_D^{r},\vspace{-6pt}
\end{equation}
where $\mathbf{g}=(g_1,\ldots,g_R)\in \mathbb{R}^{R}$, $\mathbf{U}_d\in\mathbb{R}^{I_d\times R}$ for $d=1,\ldots, D$, $||\mathbf{u}^r_{d}||_2=1$, and is the $r$-th column of matrix $\mathbf{U}_d$, and $\mathbf{u}_d^r\in\mathbb{R}^{I_d}$ for $d=1,\ldots, D$.  In other words,  $\mathcal{B}$ is the sum of $R$ rank-one tensors of dimensions $I_1,\ldots, I_D$, respectively. If $\mathcal{B}$ can be expressed exactly as the sum of $R$ rank-one tensors, then $\mathcal{B}$ admits a rank $R$ decomposition and  ``$\approx$'' in Eqn \eqref{eq:CP} can be replaced by ``$=$''. The CP decomposition can be regarded as a special case of the Tucker decomposition when the core tensor is super-diagonal and $R_d$ in  is the same for all $d=1,\ldots,D$ Eqn \eqref{eq:tucker}. If we relax the requirement of $||\mathbf{u}^r_{d}||_2=1$, $g_r$ in Eqn \eqref{eq:CP} can be incorporated in any of the $\mathbf{u}^r_{d}$ terms for $d=1,\ldots,D$. For example, if $g_r$ is multiplied with $\mathbf{u_1}$,   the CP decomposition can be written as $\sum_{r=1}^{R}\mathbf{u'}_1^{r}\circ\mathbf{u}_2^{r}\circ\dots\circ\mathbf{u}_D^{r}$, where $\mathbf{u'}_1=\mathbf{g_r}\cdot\mathbf{u}^r_{d}$ (element-wise multiplication). 

Applying  the Tucker and CP compositions, $\langle\mathcal{X}_i,\mathcal{B}\rangle$ in the TR model in Eq \eqref{eq:tucker} can be written, respectively, as\vspace{-12pt}
\begin{align}
\!\!\langle\mathcal{X}_i,\mathcal{B}\rangle&
=\langle\mathcal{X}_i, \ldbrack\mathcal{G};\mathbf{U}_1,\mathbf{U}_2,\dots,\mathbf{U}_D\rdbrack\rangle
=\langle\mathcal{X}_i, \sum_{r_1=1}^{R_1}\sum_{r_2=1}^{R_2}\dots\sum_{r_D=1}^{R_D}g_{r_1r_2\dots r_D}\mathbf{u}_1^{r_1}\circ\mathbf{u}_2^{r_2}\circ\dots\circ\mathbf{u}_D^{r_D}\rangle,\!\!\label{eq:tucker.cp}\\
\!\!\langle\mathcal{X}_i,\mathcal{B}\rangle&
= \langle\mathcal{X}_i, \ldbrack\mathbf{g};\mathbf{U}_1,\mathbf{U}_2,\dots,\mathbf{U}_D\rdbrack\rangle
=\textstyle\langle\mathcal{X}_i, \sum_{r=1}^{R}g_r\mathbf{u}_1^{r}\circ\mathbf{u}_2^{r}\circ\dots\circ\mathbf{u}_D^{r}\rangle.\!\! \label{eq:tucker.tucker}
\end{align}
\vspace{-12pt}

$\mathcal{G},\mathbf{g},\mathbf{u}$' will thus be estimated instead of the elements in the original tensor $\mathcal{B}$. As mentioned above, the model based on either of the two decompositions can lead to a potentially significant decrease in the number of parameters while honoring at least partially the structural information in $\mathcal{B}$ compared to a GLM with vectorized $\mathcal{B}$ and $\mathcal{X}$. Table \ref{tab:para} lists the numbers of free parameters in the original tensor $\mathcal{B}$, its Tucker decomposition, and its CP decomposition. For CP decomposition, the number of free parameters, compared to that in the original  $\mathcal{B}$, would be much smaller. For example, For $D=3$ and $I_d=64$, then the number of parameters goes  from $262,144$  down to  $6,080$ if $R=32$ (97.7\% decrease) and down to $12,160$ if $R=64$ (95.3\% decrease). In the case of Tucker decomposition,  $R_d$ needs to be smaller than $I_d$ at least for some $d$ in order for the number of free parameters to decrease from that in the original  $\mathcal{B}$; otherwise, the number of parameters would not be reduced though it still has the benefit of honoring the structural information in $\mathcal{X}$ via the decomposition and provides more control over the what parameters to regularize in the model.
\begin{table}[!htb]
\vspace{-6pt}\begin{center}
\caption{Number of free parameters in Tucker decomposition and CP decomposition of tensor $\mathcal{B}\in \mathbb{R}^{I_1\times I_2\times \dots \times I_D}$ \citep{li2018tucker}}\label{tab:para}\vspace{-3pt}
\resizebox{0.75\textwidth}{!}{
\begin{tabular}{lccc}
\hline
 & rank-$R$ CP & Tucker & original  \\
\hline
$D=2$ & $R(I_1+I_2)-R^2$ & $I_1 R_1+I_2R_2+R_1R_2-R_1^2-R_2^2$ &\multirow{2}{*}{$\prod_{d=1}^{D}I_d$ }\\
$D>2$ & $R(\sum_{d=1}^{D}I_d-D+1)$ & $\sum_{d=1}^{D}I_dR_d+\prod_{d=1}^{D}R_d-\sum_{d=1}^{D}R_d^2$& \\
 \hline
\end{tabular}}\vspace{-15pt}
\end{center}
\end{table}

\section{Method}\label{sec:method}\vspace{-3pt}
Though  CP decomposition is simple, it is often limited in its approximation capability and requires $\mathbf{u}_d^r$ to have the same dimension for any fixed $d$ and different $r$, where $r=1,2,\ldots,R$ in \ref{eq:CP}. In Comparison, Tucker decomposition is more flexible and allows different numbers of components in different modes after the decomposition and as well as less structure on the core tensor (CP is a special case of Tucker with a super-diagonal structure for the core tensor). For this reason, we focus on TR decomposition. Interested readers may also refer to \citep{li2018tucker} for the comparison between Tucker and CP decompositions when used in the TR setting.

We examine the case when response variable $Y$ follows an exponential family  distribution given in Eq \eqref{eq:glm}. The loss function  is the negative log-likelihood function of $\mathcal{B}$ (other loss functions without making distributional assumptions, such as the simply $\ell_2$ loss or cross-entropy loss, can be used as seen appropriate),
\begin{equation}\label{eq:loss}
l=l(\mathcal{B}|\mathbf{y},\mathcal{X}_i,1 \leqslant i \leqslant n)= \textstyle -\sum_{i=1}^{n}\ln{p(y_i|\mathcal{X}_i,\mathcal{B})}.
\end{equation}

Our proposed method \our applies Tucker decomposition to  $\mathcal{B}$ (Eq \eqref{eq:tucker}) and uses an iterative procedure to estimate the core tensor $\mathcal{G}$ with $\ell_0$ regularization and $\mathbf{U}_d$ for $d=1,\ldots, D$. Since the core tensor represents the interaction among the factor matrices 
in the Tucker decomposition of  $\mathcal{B}$, it being sparse implies that there are only limited interactions among the factor matrices, which helps with interpretations.

Each iteration $t$ in the \our algorithm contains two simple steps: 1) generate noisy data ($\mathcal{Z}^{(t)}_j, e^{(t)}_{y,j}$) for $j=1,\ldots,n_e$ given the updated tensor estimate from iteration $t-1$; 2) run GLM on the updated noise-augmented data 
\begin{equation}\label{eq:data}
\left[\begin{matrix}
 (\mathcal{X}_i, y_i)_{i=1,\ldots,n}\\
 (\mathcal{Z}^{(t)}_j, e^{(t)}_{y,j})_{j=1,\ldots,n_e}\\
 \end{matrix}\right].
 \end{equation}
Values of noisy response data $e^{(t)}_{y,j}$ depend on the type of TR.  For linear TR, we may set $e^{(t)}_{y,j}\equiv0$, the sample mean $\bar{y}$, or any other constant $d$ for $j=1,\ldots,n_e$; for logistic TR with $Y=\{0,1\}$, we may set half of $n_e$ $e^{(t)}_{y,j}$'s at 0 and the other half at 1; for Poisson and  negative binomial TR, we may set $e^{(t)}_{y,j}=1$ for $j=1,\ldots,n_e$. Augmented tensor predictor $\mathcal{Z}^{(t)}_j$ is constructed from the components of a Tucker decomposition; that is, 
\begin{align}
&\mathcal{Z}^{(t)}_j \triangleq  \mathcal{E}^{(t)}_j\times_1\mathbf{U}^{(t-1)}_1\times_2\mathbf{U}^{(t-1)}_2\times_3\dots\times_\mathrm{D}\mathbf{U}^{(t)-1}_D\notag\\
=&\textstyle \sum_{r_1=1}^{R_1}\sum_{r_2=1}^{R_2}\dots\sum_{r_D=1}^{R_D} 
e^{(t)}_{j,i_1i_2\dots i_D} \mathbf{u}_1^{r_1(t-1)}\circ\mathbf{u}_2^{r_2(t-1)}\circ\dots\circ\mathbf{u}_D^{r_D(t-1)},
\label{eq:zj}\\
\mbox{where }&e^{(t)}_{j,i_1i_2\dots i_D}\sim\mathcal{N}\big(0,\lambda \left(g^{(t-1)}_{r_1r_2\dots r_D}\right)^{-2}\big),\label{eq:e}\
\end{align}
$\lambda$ is a hyperparameter (tuned or prespecified), $g^{(t-1)}_{r_1r_2\dots r_D}$ is the element $[r_1,r_2,\dots r_D]$ in the core tensor $\mathcal{G}^{(t-1)}$, and $\mathbf{U}^{(t-1)}_i$ for $i=1,\ldots,D$ is the estimate of $\mathbf{U}_i$ from iteration $t-1$; both $\mathcal{G}^{(t-1)}$ and $\mathbf{U}^{(t-1)}_i$  are the components in the Tucker decomposition of estimate $\mathcal{B}^{(t-1)}$.  Eq \eqref{eq:e} is referred to as the \emph{noise-generating distribution}, a Gaussian distribution with an adaptive variance term that is  designed to achieve $\ell_0$ regularization. In addition, $n_e$ needs to be $<p$, where $p$ is the total number of elements in $\mathbf{G}$, to achieve the $\ell_0$ regularization (the number of zero elements in theory in $\mathbf{G}$ is  $n_e$; established in Sections \ref{sec:linear}).

After the noisy data are generated, we run a regular GLM to obtain an updated estimate  $\mathcal{B}^{(t)}$ by minimizing the loss function formulated given the noise-augmented data, which is
\begin{align} \label{eq:loss.na}
l^{(t)}_{na}&=l(\mathcal{B},\phi|\mathbf{y}_i,\mathcal{X}_i,e^{(t)}_{y,j},\mathcal{Z}^{(t)}_j,1 \leqslant i \leqslant n,1 \leqslant j \leqslant n_e)\notag\\
&\textstyle=-\sum_{i=1}^{n}\ln{p(y_i|\mathcal{X}_i,\mathcal{B},\phi)}-\sum_{j=1}^{n_e}\ln{p(e^{(t)}_{y,j}|\mathcal{Z}^{(t)}_j,\mathcal{B},\phi)}.
\end{align}
This step can leverage any software that runs regular GLM (without regularization) without a need to design or code an optimization procedure  to estimate $\mathcal{B}$.

In what follows, we establish theoretically that \our achieves $\ell_0$ regularization on the core tensor $\mathcal{G}$ upon convergence and provide some intuition on how that is achieved. We first examine linear TR with Gaussian $Y$ (Section \ref{sec:linear}) and then extend the conclusion to GLM TR in general (Section \ref{sec:gl}). 

\vspace{-3pt}\subsection{\texorpdfstring{$\ell_0$}{} regularization via \texorpdfstring{\our}{} in Linear TR}\label{sec:linear}
In  linear TR, the loss function is the $\ell_2$ loss.  Given  observed data $(\mathcal{X}_i,y_i)$) for $i=1,\ldots,n$ and noisy data ($\mathcal{Z}_i,e_{y,j}\equiv d)$  for $j=1,\ldots,n_e$ that augment the former, the loss function is
\begin{align} \label{eq:loss.na.linear0}
l_{na}&=\textstyle\sum_{i=1}^{n}(y_i-\langle\mathcal{X}_i,\mathcal{B}\rangle)^2 +\sum_{j=1}^{n_e}(e_{y,j}-\langle\mathcal{Z}_i,\mathcal{B}\rangle)^2\notag\\
&=\textstyle\sum_{i=1}^{n}(y_i-\langle\mathcal{X}_i,\mathcal{B}\rangle)^2 +n_ed^2-2d\sum_{j=1}^{n_e}\langle\mathcal{Z}_j,\mathcal{B}\rangle+
\sum_{j=1}^{n_e}\langle\mathcal{Z}_j,\mathcal{B}\rangle)^2.
\end{align}
Eq \eqref{eq:loss.na.linear0} contains both a linear term and a quadratic term  in $\mathcal{B}$. We show below that it is the quadratic term that yields $\ell_0$ regularization. Toward that end, augmented noisy data need to be designed in a way such that the linear term is zero so to avoid any undesirable regularization effect on $\mathcal{B}$ from the linear term. In the linear TR case, there are two ways to achieve the goal. First, we can set $e_{y,j}\equiv0$  for all $j=1,\ldots,n_e$, in which case, the linear term  $2d\sum_{j=1}^{n_e}\langle\mathcal{Z}_j,\mathcal{B}\rangle)=0$ in Eq \eqref{eq:loss.na.linear0}. Second, we  generate two blocks of noisy terms, each of size $n_e$. The first $n_e$ terms contain noisy data points $(e_{y,j},\mathcal{Z}_j)$ for $1 \leqslant j \leqslant n_e$ and the second  block contains data points $(e_{y,j}=e_{y,j-n_e},\mathcal{Z}_j=-\mathcal{Z}_{j-n_e})$ for $n_e+1 \leqslant j \leqslant 2n_e$, as demonstrated in Eq \eqref{eq:data2},
\begin{equation}\label{eq:data2}
\left[\begin{matrix}
(\mathcal{X}_i, y_i)_{i=1,\ldots,n}\\
(\mathcal{Z}^{(t)}_j, e^{(t)}_{y,j})_{j=1,\ldots,2n_e}\\
\end{matrix}\right]=
\left[\begin{matrix}
(\mathcal{X}_i, y_i)_{i=1,\ldots,n}\\
(\mathcal{Z}'^{(t)}_j, e'^{(t)}_{y,j})_{j=1,\ldots,n_e}\\
(-\mathcal{Z}'^{(t)}_j, e'^{(t)}_{y,j})_{j=1,\ldots,n_e}\\
\end{matrix}\right].
\end{equation}
The linear term in Eq \eqref{eq:loss.na.linear0} is also 0  as $\sum_{j=1}^{n_e}\langle\mathcal{Z}_j,\mathcal{B}\rangle+ \sum_{j=n_e+1}^{2n_e}\langle\mathcal{Z}_j,\mathcal{B}\rangle=0$ per the design in Eq \eqref{eq:data2}. In summary, both noise augmentation schemes lead to 
\begin{align} 
l_{na}&=\textstyle\sum_{i=1}^{n}(y_i-\langle\mathcal{X}_i,\mathcal{B}\rangle)^2 +C_1\sum_{j=1}^{n_e}\langle\mathcal{Z}_j,\mathcal{B}\rangle^2+C_2,
\label{eq:loss.na.linear}
\end{align}
where $C_1=1,C_2=0$  if we set $e_{y,j}\equiv0$ and $C_1=2, C_2=2n_ed^2$ if we use the two-block noise augmentation scheme. 

We now show the quadratic term $\sum_{j=1}^{n_e}\langle\mathcal{Z}_j,\mathcal{B}\rangle^2$ in Eq \eqref{eq:loss.na.linear} leads to $\ell_0$ regularization on the core tensor of $\mathcal{B}$ if Tucker decomposition is applied to $\mathcal{Z}$ and $\mathcal{B}$.  The inner product $\langle\mathcal{Z}_j,\mathcal{B}\rangle$ in Eq \eqref{eq:loss.na.linear}, after Tucker decomposition on  $\mathcal{B}$ and given how $\mathcal{Z}$ is designed, can be written as
\begin{align}
&\bigg\langle\sum_{i_1=1}^{R_1}\!\sum_{i_2=1}^{R_2}\!\dots\!\sum_{i_D=1}^{R_D}\!e_{j,i_1i_2\dots i_D}\mathbf{u}_1^{i_1}\circ\mathbf{u}_2^{i_2}\circ\dots\circ\mathbf{u}_D^{i_D},
\sum_{i_1=1}^{R_1}\!\sum_{i_2=1}^{R_2}\!\dots\!\sum_{i_D=1}^{R_D}\!g_{i_1i_2\dots i_D}\mathbf{u}_1^{i_1}\circ\mathbf{u}_2^{i_2}\circ\dots\circ\mathbf{u}_D^{i_D}\bigg\rangle\notag\\
=&\sum_{i_1=1}^{R_1}\sum_{i_2=1}^{R_2}\dots\sum_{i_D=1}^{R_D}\langle e_{j,i_1i_2\dots i_D}\mathbf{u}_1^{i_1}\circ\mathbf{u}_2^{i_2}\circ\dots\circ\mathbf{u}_D^{i_D},g_{i_1i_2\dots i_D}\mathbf{u}_1^{i_1}\circ\mathbf{u}_2^{i_2}\circ\dots\circ\mathbf{u}_D^{i_D}\rangle+\notag\\
&\sum_{i_d\neq j_d,1\leqslant i_d,j_d \leqslant R_d}\langle e_{j,i_1i_2\dots i_D}\mathbf{u}_1^{i_1}\circ\mathbf{u}_2^{i_2}\circ\dots\circ\mathbf{u}_D^{i_D},g_{j_1ij_2\dots j_D}\mathbf{u}_1^{j_1}\circ\mathbf{u}_2^{j_2}\circ\dots\circ\mathbf{u}_D^{j_D}\rangle\label{eq:ZjB0}\\
=&\textstyle\sum_{i_1=1}^{R_1}\sum_{i_2=1}^{R_2}\dots\sum_{i_D=1}^{R_D} e_{j,i_1i_2\dots i_D}g_{i_1i_2\dots i_D}
\label{eq:ZjB},
\end{align}

Eq \eqref{eq:ZjB} holds due to the orthonormality of the columns in $\mathbf{u}_d$ (thus $\langle \mathbf{u}_1^{i_1}\circ\mathbf{u}_2^{i_2}\circ\dots\circ\mathbf{u}_D^{i_D},\mathbf{u}_1^{i_1}\circ\mathbf{u}_2^{i_2}\circ\dots\circ\mathbf{u}_D^{i_D}\rangle=1$ in the first term and the second term in Eq \eqref{eq:ZjB0}  is zero). 
Substituting $\langle\mathcal{Z}_j,\mathcal{B}\rangle$  in Eq \eqref{eq:loss.na.linear} with Eq \eqref{eq:ZjB}, we have 
\begin{align}
l_{na}=&\sum_{i=1}^{n}(y_i-\langle\mathcal{X}_i,\mathcal{B}\rangle)^2 +C_1\sum_{j=1}^{n_e}\bigg(\sum_{i_1=1}^{R_1}\sum_{i_2=1}^{R_2}\dots\sum_{i_D=1}^{R_D}g_{i_1i_2\dots i_D}e_{j,i_1i_2\dots i_D}\bigg)^2+C_2\\ 
=&\sum_{i=1}^{n}(y_i-\langle\mathcal{X}_i,\mathcal{B}\rangle)^2 +\lambda\sum_{j=1}^{n_e}\bigg(\sum_{i_1=1}^{R_1}\sum_{i_2=1}^{R_2}\dots\sum_{i_D=1}^{R_D}g_{i_1i_2\dots i_D}e'_{j,i_1i_2\dots i_D}\bigg)^2+C_2,\label{eq:loss.na.linear1}\\  
\mbox{where } & e'_{j,i_1i_2\dots i_D}=\lambda^{-1/2}e_{j,i_1i_2\dots i_D} \sim\mathcal{N}(0,g^{-2}_{i_1i_2\dots i_D}).\label{eq:e'}
\end{align}
Eq \eqref{eq:loss.na.linear1} suggests that the second term of the loss function formulated on augmented noisy data serves as a regularization term on the core tensor $\mathcal{G}$ of $\mathcal{B}$.  

The key to understand  methodologically and intuitively why  \our with the carefully constructed data augmentation scheme achieves $\ell_0$ regularization is to realize that the optimization problem in Eq \eqref{eq:loss.na.linear1} is the dual problem to and the  Lagrange expression of the  primal constrained optimization problem in Eq \eqref{eq:constraint1} with the same multiplier $\lambda$ used for all $n_e$ constrained terms; 
\begin{align}
&\textstyle \min_{\mathcal{B}}\sum_{i=1}^{n}(y_i-\langle\mathcal{X}_i,\mathcal{B}\rangle)^2, \mbox{ subject to}\notag\\
&\textstyle \sum_{i_1=1}^{R_1}\sum_{i_2=1}^{R_2}\dots\sum_{i_D=1}^{R_D}g_{i_1i_2\dots i_D}e'_{1,i_1i_2\dots i_D}=0\notag\\
&\textstyle\sum_{i_1=1}^{R_1}\sum_{i_2=1}^{R_2}\dots\sum_{i_D=1}^{R_D}g_{i_1i_2\dots i_D}e'_{2,i_1i_2\dots i_D}=0,\label{eq:constraint1} \\
&\qquad\vdots\notag\\
&\textstyle\sum_{i_1=n_e}^{R_1}\sum_{i_2=1}^{R_2}\dots\sum_{i_D=1}^{R_D}g_{i_1i_2\dots i_D}e'_{n_e,i_1i_2\dots i_D}=0.\notag
\end{align}
Eq \eqref{eq:constraint1} shows that the $n_e$ linear constraints in the primal problem promote orthogonality between the core tensor elements $g$'s and augmented noisy tensor predictors. With $n_e< p$, the $n_e$ constraints only affect a subset of the $p$ parameters. In each iteration  of the \our algorithm with a new set of $n_e$ noisy samples, the fix subset comprising of the $n$ observed samples is responsible for learning the relationship (the GLM coefficients) between the predictors $\mathbf{X}$ and the response $Y$ while the noisy subset drives the estimates of parameters $g$'s to be orthogonal to the hyper plane spanned by the $n_e$ noisy predictor vectors because that's when the loss constructed on the noisy data subset in Eq \eqref{eq:loss.na.linear1} is minimized.  
Per the design of the noise generation distribution, the smaller magnitude a $g$ is of, the larger the variance of the noise-generating distribution, and the more dispersed the  noisy predictor corresponding to that $g$ is and the less relevant  it is for the prediction of $Y$ in the TR model, further pushing the estimated value of $g$ towards zero, until stabilization and convergence. 

Proposition \ref{prop:l0} states the constrained optimization Eqn \eqref{eq:constraint1} is $\ell_0$ regularization with exactly $n_e$ elements in $\mathcal{G}$  set at 0.  The proof is straightforward. There are $n_e$ linear constraints on the elements $g$'s of in $\mathcal{G}$ n Eqn \eqref{eq:constraint1}, the coefficients of which are noise terms drawn independently from the noise-generating distribution, implying that exactly $n_e$ entries  can be expressed exactly as a linear function of the rest of the components in $\mathcal{G}$; in other words, $n_e$ components in the core tensor can be set at 0. 
\begin{prop}[$\ell_0$ regularization on core tensor in linear TR through \our] \label{prop:l0}
The constrained optimization problem in Eqn \eqref{eq:constraint1} is equivalent to
\begin{align}
&\textstyle \min_{\mathcal{B}}\sum_{i=1}^{n}(y_i-\langle\mathcal{X}_i,\mathcal{B}\rangle)^2\notag\\
\mbox{subject to }&\textstyle \sum_{i_1=1}^{R_1}\sum_{i_2=1}^{R_2}\dots\sum_{i_D=1}^{R_D}\mathbbm{1}(g_{i,i_1i_2\dots i_D}= 0)= n_e.\label{eq:constraint2}
\end{align}
\end{prop}

Since $g_{i_1i_2\dots i_D}$ is unknown, the iterative \our procedure would require users to supply an initial $\mathcal{B}^{(0)}$ and then use the estimates of $\mathcal{G}$ from the previous iteration when sampling noisy tensor predictor data in Eq \eqref{eq:e}. Upon convergence, the estimate of $\mathcal{G}$ stabilizes, and so is the noise-generating distribution; but the random fluctuation in the estimate of $\mathcal{G}$ still exists from iteration to iteration given that randomly sampled noisy data are different across the iterations.  For that reason, while the $\ell_0$ regularization via \our is exact conceptually, the realized regularization in actual implementations of \our would only get arbitrarily close to $\ell_0$ due to the random fluctuation.  One way to mitigate the numerical randomness around the estimate of $\mathcal{B}$ so to  get as close as possible to exact $\ell_0$ is to take the average of estimated parameters over multiple iterations upon convergence; details are provided in Algorithm \ref{al1} of Section \ref{sec:algorithm}.

\subsection{\texorpdfstring{$\ell_0$}{} regularization via \texorpdfstring{\our}{} in GLM TR}
\label{sec:gl}
When $Y$ follows an exponential family distribution in Eq \eqref{eq:glm}, the loss function can be formulated as a negative log-likelihood. We employ the same noisy-data augmentation scheme as  in Eq \eqref{eq:data2} with the two blocks of noisy data, each of size $n_e$ with opposite signs on the noisy predictor tensor when augmenting the loss function for NA$_0$CT$^2$; that is, 
\begin{align}\label{eq:lna}
&l_{na}(\mathcal{B},\phi|y_i,\mathcal{X}_i,e_{y,j},\mathcal{Z}_j,1\!\leqslant \!i \!\leqslant\! n,1\! \leqslant\! j \leqslant \!2n_e)\notag\\
&= -\!\sum_{i=1}^{n}\ln{p(y_i|\mathcal{X}_i,\mathcal{B},\phi)}-\! \sum_{j=1}^{n_e}\ln{p(e_{y,j}|\mathcal{Z}_j,\mathcal{B},\phi)}-\! \sum_{j=n_e+1}^{2n_e}\ln{p(e_{y,j}|-\mathcal{Z}_j,\mathcal{B},\phi)}.
\end{align}
\begin{prop}[$\ell_0$ regularization on core tensor in general TR through \our] \label{prop:l0glm}
Minimization of the augment loss function in Eqn \eqref{eq:lna} is equivalent to
\begin{align}
&\textstyle \min_{\mathcal{B}}-\!\sum_{i=1}^{n}\ln{p(y_i|\mathcal{X}_i,\mathcal{B},\phi)}\notag\\
\mbox{subject to }&\textstyle \sum_{i_1=1}^{R_1}\sum_{i_2=1}^{R_2}\dots\sum_{i_D=1}^{R_D}\mathbbm{1}(g_{i,i_1i_2\dots i_D}= 0)= n_e.\label{eq:constraint3}
\end{align}
\end{prop}

Propositions \ref{prop:l0glm} is proved as follows.  Applying the Taylor expansion around  $\langle\mathcal{Z}_j,\mathcal{B}\rangle=0$,
$l_{na}$ in Eq \eqref{eq:lna} becomes
\begin{align}
&-\textstyle\sum_{i=1}^{n}\ln{p(y_i|\mathcal{X}_i,\mathcal{B},\phi)}
-\textstyle\sum_{j=1}^{n_e}\!\bigg\{\!\frac{-B(0)}{a(\phi)}+h(e_{y,j},\phi)+
\cancel{\frac{e_{y,j}-B^{\prime}(0)}{a(\phi)}\langle\mathcal{Z}_j,\mathcal{B}\rangle}+
\frac{B^{\prime\prime}(0)}{2a(\phi)}\langle\mathcal{Z}_j,\mathcal{B}\rangle^2 \notag\\
&+\cancel{O\big(\langle\mathcal{Z}_j,\mathcal{B}\rangle^3\big)}
+O\big(\langle\mathcal{Z}_j,\mathcal{B}\rangle^4\big)+\ldots\!\bigg\}-\textstyle\sum_{j=n_e+1}^{2n_e}\!\bigg\{\!\frac{-B(0)}{a(\phi)}+h(e_{y,j},\phi)+
\cancel{\frac{e_{y,j}-B^{\prime}(0)}{a(\phi)}\langle 
-\mathcal{Z}_j,\mathcal{B}\rangle}\notag\\
&+\frac{B^{\prime\prime}(0)}{2a(\phi)}\langle-\mathcal{Z}_j,\mathcal{B}\rangle^2
+\cancel{O\big(\langle-\mathcal{Z}_j,\mathcal{B}\rangle^3\big)}
+O\big(\langle-\mathcal{Z}_j,\mathcal{B}\rangle^4\big)+\ldots\!\bigg\}\notag\\
=&\textstyle\!-\!\sum_{i=1}^{n}\ln{p(y_i|\mathcal{X}_i,\mathcal{B},\phi)} 
\!+\!C_1\sum_{j=1}^{n_e}\sum_{l=1}^\infty\langle\mathcal{Z}_j,\mathcal{B}\rangle^{2l}\!+C_2,
\label{eq:loss.glm}
\end{align}
where  $C_1=-B^{\prime\prime}(0)/(a(\phi))$ and $C_2=2n_eB(0)/a(\phi)-\sum_{j=1}^{2n_e}h(e_{y,j},\phi)$ are constant independent of $\mathcal{B}$. The augmented loss function in  Eq \eqref{eq:loss.glm} compared to the the regularization term in Eq \eqref{eq:loss.na.linear} in linear TR, has additional even exponents of $\langle\mathcal{Z}_j,\mathcal{B}\rangle$ in addition to the quadratic term $\langle\mathcal{Z}_j,\mathcal{B}\rangle^{2l}$. These additional even-powered terms of $\langle\mathcal{Z}_j,\mathcal{B}\rangle$ have the same regularization effect as the quadratic term, which is orthogonality constraints between $\mathbf{e}_j$ and the core tensor $\mathcal{G}$ or the $\ell_0$ regularization on  $\mathcal{G}$  just as in the case of linear TR.  Therefore, the statements from Propositions \ref{prop:l0}  also apply to the GLM TR setting, 

\vspace{-6pt}\subsection{Algorithm for \texorpdfstring{\our}{}}\label{sec:algorithm}\vspace{-3pt}
\small
\setstretch{1}
\begin{algorithm}[H]
\caption{The \our procedure}\label{al1}
\SetAlgoLined
\SetKwInOut{Input}{input}
\SetKwInOut{Output}{output}
\Input{observed data $(\mathcal{X}_i,y_i)$ for $i=1,\ldots,n$; noisy data size $n_e$,  noisy response data $e_{y,j}$ for $j=1,\ldots,n_e$; regularization hyperparameter $\lambda$; maximum iterations $T$;  stopping thresholds $\tau$ or $\eta$; zero threshold $\tau_0$,  moving average window $m$; dimension of core tensor $\mathcal{G}: R_1\times R_2\times\cdots\times R_D$ if applicable (optional)}
\Output{TR coefficient tensor $\hat{\mathcal{B}}$}
Standardize $\mathcal{X}_i$ for each $1\leqslant i\leqslant n$\;
Calculate $\hat{\mathcal{B}}$ from regression model on $Y$ on vectorized $\mathcal{X}_i$ and set $\bar{\mathcal{B}}^{(0)}=\hat{\mathcal{B}}^{(0)}=\hat{\mathcal{B}}$\;
 $t\leftarrow 1$, convergence status $s\leftarrow 0$\;
\While{$t\leqslant T$ and $s=0$}
{
  Apply Tucker decomposition to $\hat{\mathcal{B}}^{(t-1)}=\hat{\mathcal{G}}^{(t-1)}\times_1\hat{\mathbf{U}}^{(t-1)}_1\times_2 \hat{\mathbf{U}}^{(t-1)}_2\times_3\dots \times_\mathrm{D}\hat{\mathbf{U}}^{(t-1)}_D$\;
  \eIf{$t\leqslant m$}{
   $\bar{\mathcal{G}}^{(t-1)}\leftarrow\hat{\mathcal{G}}^{(t-1)}$\;
   }
   {$\bar{\mathcal{G}}^{(t-1)}\leftarrow\big\{\sum_{i=t-m-1}^t\hat{\mathcal{G}}^{(i)}/m+c\}$\; \tcp{$c$ is a small constant (e.g., $10^{-7}$) to avoid overflow of $(\bar{\mathcal{G}}^{(t-1)})^{-2}$.}
   }
   Generate elements in the core tensor $\mathcal{E}^{(t)}_j$ of the noisy predictor $\mathcal{Z}_j$: $e_{j,i_1i_2\dots i_D}\sim \mbox{N}\big(0,\lambda\big(\bar{g}_{i_1i_2\dots i_D}^{(t-1)}\big)^{-2}\big)$ for $j=1,\dots,n_e$, $i_d=1,\ldots,R_d$, and $d=1,\ldots,D$\;
   Calculate $\mathcal{Z}_j=\mathcal{E}^{(t)}_j\times_1\hat{\mathbf{U}}^{(t-1)}_1\times_2\hat{\mathbf{U}}^{(t-1)}_2\times_3\dots\times_\mathrm{D}\hat{\mathbf{U}}^{(t-1)}_D$ for $j=1,\dots,n_e$\;
   Combine $(\mathcal{X}_i,y_i)_{i=1,\ldots,n}$, $(\mathcal{Z}^{(t)}_j,e_{y,j})_{j=1,\ldots,n_e}$, and $(-\mathcal{Z}^{(t)}_j,e_{y,j})_{j=1,\ldots,n_e}$  to form an augmented dataset of size $n+2n_e$ \;

   Run GLM on the augmented data with vectorized predictor vec$(\mathcal{X}_i,\mathcal{Z}^{(t)}_j)_{i=1,\ldots,n;j=1,\ldots,2n_e}$ and response $(y_i,e_{y,j})_{i=1,\ldots,n;j=1,\ldots,2n_e}$ to obtain updated estimate $\hat{\mathcal{B}}^{(t)}$\;
   \eIf{$t\leqslant m$}{
   $\bar{\mathcal{B}}^{(t)}\leftarrow\hat{\mathcal{B}}^{(t)}$ and  $\bar{l}^{(t)}\leftarrow\hat{l}(\hat{\mathcal{B}}^{(t)})$\;
   }{$\bar{\mathcal{B}}^{(t)}\leftarrow\sum_{i=t-m-1}^t\hat{\mathcal{B}}^{(i)}/m$ and $\bar{l}^{(t)}\leftarrow\sum_{i=t-m-1}^t\hat{l}(\hat{\mathcal{B}}^{(i)})/m$\;
   }
   \If{$|\bar{l}^{(t)}-\bar{l}^{(t-1)}|\leqslant \tau$ or $||\bar{\mathcal{B}}^{(t)}-\bar{\mathcal{B}}^{(t-1)}||_1\leqslant\eta$}{$s=1$\;}
  $t\leftarrow t+1$\;
}
Output $\bar{\mathcal{B}}^{(t)}$.\tcp{the next two steps are optional}
Apply Tucker decomposition to $\bar{\mathcal{B}}^{(t)}$ to obtain $\hat{\mathcal{G}}\times_1\hat{\mathbf{U}}_1\times_2\hat{\mathbf{U}}_2\times_3\dots\times_\mathrm{D}\hat{\mathbf{U}}_D$, replace any element  in  $\hat{\mathcal{G}}$ that is $\le\tau_0$  with 0 to obtain a new $\Hat{}{\mathcal{G}}_0$\;
Output $\hat{\mathcal{B}}=\hat{\mathcal{G}}_0\times_1\hat{\mathbf{U}}_1\times_2\hat{\mathbf{U}}_2\times_3\dots\times_\mathrm{D}\hat{\mathbf{U}}_D$.
\end{algorithm}

\normalsize
Algorithm \ref{al1} lists the steps for running the \our procedure. The algorithm uses an  explicitly specified stopping criterion. Besides that, plotting trace plots of loss functions to eyeball the convergence   or examine the change in the estimate of $\hat{\mathcal{B}}$ across iterations such as $||\bar{\mathcal{B}}^{(t)}-\bar{\mathcal{B}}^{(t-1)}||_1\leqslant \eta$, where $\eta$ is a small constant, can also be used to evaluate convergence.  Users need to supply an initial set of parameters, such as from the regression model on vectorized $\mathcal{X}_i$, which can be regularized or not, as given in line 2. The algorithm also includes a moving window $m>1$ to help to smooth out the randomness in the estimated tensor $\hat{\mathcal{B}}$ and in the loss function since augmented noisy data change at every iteration. The usage of $m$ is  algorithmic and doesn't affect the theoretical property of \our established in Proposition \ref{prop:l0}.   Since the variances of noise-generating Gaussian distributions are inversely proportional to the squared core tensor entry values, the estimates of zero-valued entries in the core tensor via \our would   shrink toward 0 upon the convergence of the algorithm, 

leading to very large variances in the noise-generating distributions and potential  overflows when running the algorithm. To avoid this, we recommend lower bounding the estimated elements in the core tensor by a very small constant $\tau_0$ (e.g, $10^{-10}$).

\vspace{-6pt} \subsubsection{Choice of hyperparameters} 
The two hyperparameters in the loss function of \our $\lambda$ and $n_e$ both affect the number of estimated zeros in the core tensor and can be tuned via cross-validation (CV) with metrics such as information criteria (e.g., AIC) from  GLMs. First, to achieve the $\ell_0$ regularization on the core tensor, $n_e$ should be set at a value smaller than $p$, the product of the dimensionalities of the core tensor. Second, our empirical studies suggest that 1) small $\lambda$ may not yield enough sparsity even when $n_e$ is larger than the true number of zero $n_e$ and close to $p$; 2) when $\lambda$ is large, the  estimated number of zeros in the core tensor is roughly equal to the value of $n_e$ when $n_e\le n_0$ and increases with $n_e$ for $n_e>n_0$; 3) when $\lambda$ is not small or too large,  the  estimated number of zeroes in the core tensor increases with $n_e$ and reaches a plateau valued at $n_0$. In other words, with an appropriate chosen $\lambda$, the estimated number of zeros would be robust to the choice of $n_e$ and equals to $n_0$ (see Appendix B for an empirical demonstration).
 
Before providing a recommendation on the choice $m$, we first provide the rationale for having a moving window of size $m$. Algorithmically, averaging over $m$ iterations helps smooth the parameter estimates,  improving and accelerating the the algorithm's convergence, especially considering every iterations is based on a different augmented dataset. More importantly, $m$ is more than just an algorithmic parameter to define a smoothing window.   As shown in \citet{li2022adaptive},  theoretically, it would  require  $m\to\infty$ to achieve the almost sure (a.~s.~) convergence of  loss function given the noise-augmented data to its expectation and the a.~s.~convergence of the minimizer of the former to the minimizer of the latter when fixed $n_e\le p$. This implies that in practical application of this algorithm, a large number of $m$ would be needed. Besides, the noisy-data augmentation regularization with a fixed $n_e$ also exhibits ensemble learning behavior upon convergence,  where the model trained on each of the $m$ augmented datasets is a weak learner, generating the diversity among the final estimates of the model parameters, which are averages over the $m$ sets. 
With an understanding of the functionality of $m$, we recommend selecting a relatively large value of $m$, factoring in the complexity of the problem. We run some experiments to examine how $m$ affects the convergence of the \our algorithm and the results are presented in Appendix A. As expected, a larger $m$ would lead to better convergence in terms of more stabilized loss function with less fluctuation and smaller loss values given its implicit ensemble learning behavior.
We may also adaptively tune $m$ after the algorithm starts by monitoring the residual deviance. If noticeable divergence or fluctuations occurs with the preset $m$, increasing it can help achieve stable convergence. 

The input of the dimension of $\mathcal{G}$ in Algorithm \ref{al1} is optional and  depends on whether there is prior information to inform the sparsity and dimensionality of $\mathcal{G}$ relative to $\mathcal{B}$. If no such prior information exists, dimension of $\mathcal{G}$ is set to be the same  as  $\mathcal{B}$ by default.   If there is prior information on the dimensions of one or more modes of $\mathcal{B}$, one may set that mode at the smaller dimension(s) accordingly, incorporating the prior knowledge and also helping reduce the computational cost later on when running the algorithm. 

Note that setting the $\mathcal{G}$ at a smaller dimension than $\mathcal{B}$, running tensor regression without regularization,  and constructing $\mathcal{B}$  per Eqn \eqref{eq:tucker} (referred as \emph{smaller-core-tensor without regularization} or \emph{SCOT-WOR} for brevity) does not have quite the same  sparsity  effect as imposing the $\ell_0$ constraint via Algorithm \ref{al1}, though there is some overlap. Specifically, there are a couple of key differences between the two approaches. First, \emph{SCOT-WOR} would need to tune/specify $D$ hyperparameters (i.e., $R_1\times\ldots\times R_D$) whereas \our only tune hyperparameters $\lambda$ and $n_e$. If grid search is used to tune  $R_1,\ldots,R_D$ in SCOT-WOR, it can be computationally costly as there  are $\prod_{d=1}^D k_d$ possible scenarios to choose from, where $k_d$ is the dimensionality of the grid for $R_d$. In contrast, \our\hspace{-4pt} has only two  hyperparameters $n_e$ and $\lambda$ and it is robust to the value of $n_e$ in that the correct number ($n_0$) of structural 0's in $\mathcal{G}$  can be learned for a wide range of $n_e\in[n_0,p]$ for a given proper $\lambda$ value. Second, by pre-specifying  $R_1,\ldots,R_D$, the structure of  $\mathcal{G}$ is set, lacking the flexibility that \our offers to $\mathcal{G}$ on its structure -- that is, the locations of the non-zero elements in $\mathcal{G}$ is completely data driven without following a pre-specified pattern. In some sense, \our can be viewed as naturally incorporating SCOT-WOR because $R_d\le I_d$ in Eqn \eqref{eq:tucker}, where $I_1\times\ldots\times I_D$ is the dimensionality of $\mathcal{B}$. In practice, one can set $\mathcal{G}$ with $R_d\le I_d$ for $d=1,\ldots,D$ in a confident but conservative manner (e.g., if $I_1=50$, $R_1=45$ based on some prior knowledge), and then run Algorithm \ref{al1} to impose further sparsity constraints on $\mathcal{G}$. This approach is not only more computationally efficient with a smaller $\mathcal{G}$ to estimate but also leverages the flexibility the noise-augmented $\ell_0$ regularization \our offers to its structure.

\subsubsection{Time complexity} 
The time complexity of Algorithm 1 can be analyzed by breaking down the major steps and examining the time complexity of each step, as presented in Table \ref{tab:time}. In summary, the time complexity is cubic in the dimensionality $\mathcal{B}$ ($I_1\times\dots\times I_D$) and  $\mathcal{G}$ ($R_1\times\dots\times R_D$)  and linear in $T\cdot (n+n_e)$.
\vspace{-6pt}\begin{table}[!htb]
\centering
\caption{Time complexity of the key steps in Algorithm \ref{al1}} \label{tab:time}\vspace{-3pt}
\resizebox{0.9\textwidth}{!}{
\begin{tabular}{l|l|l}
\hline
& line in Algorithm \ref{al1} & time complexity\\
\hline
every & 5: Tucker decomposition of $\hat{\mathcal{B}}^{(t-1)}$ & $O\!\left(\sum_{t=1}^T\!T^{(t-1)}_{\text{tucker}}\!\left(\!pD \!+\! \sum_{d=1}^{D}\!I_d R_d \prod_{j \neq d} R_j\!\right)\!\right)^{\dagger}$ \\
iteration & 12: Calculation of $\mathcal{Z}_j$ &  $O\!\left(Tn_e\sum_{j=1}^D I_j R_j \prod_{j' \neq d} R_{j'} \right)^{\ddagger}$\\
$t$ & 14: GLM on the augmented data & $O(\sum_{t=1}^TT^{(t)}_{\text{glm}}(p^2(n+2n_e) + p^3))^{\ddagger}$\\
\hline
one &26: Tucker decomposition of $\bar{\mathcal{B}}^{(t)}$ & $O\left(T'_{\text{tucker}}\!\left(\!pD \!+\! \sum_{d=1}^D\!I_d R_d \prod_{j \neq d} R_j\! \right)\right)^{\dagger}$\\
time & 27: calculation of $\hat{\mathcal{B}}$ & $O\!\left(\!\sum_{j=1}^D I_j R_j \prod_{j' \neq d} R_{j'}\!\right)^{\ddagger}$\\
\hline
\end{tabular}}
\resizebox{0.9\textwidth}{!}{
\begin{tabular}{l@{}l}
\small{total}:& $\; O\!
\left(\left(\sum_{t=1}^T\!T^{(t-1)}_{\text{tucker}}\!+\!T'_{\text{tucker}}\right)\!\cdot\left(\!pD \!+\! \sum_{d=1}^{D}\!I_d R_d \prod_{j \neq d} R_j\!\right)\!\right)\!+\!O\!\left((Tn_e+1)\sum_{j=1}^D I_j R_j \prod_{j' \neq d} R_{j'} \right)$\\
& $+ O\big(\sum_{t=1}^TT^{(t)}_{\text{glm}}(p^2(n+2n_e) + p^3)\big)$, where  $p=\prod_{d=1}^D I_d$.\\
\hline
\end{tabular}}
\resizebox{0.9\textwidth}{!}{
\begin{tabular}{l}
$^{\dagger}$ based on the Higher-Order Orthogonal Iteration (HOOI) algorithm for Tucker decomposition \\
(e.g., used by the \texttt{tucker} function in \texttt{rTensor} in R). $T^{(t-1)}_{\text{tucker}}$ is the number  of iterations  for HOOI \\
to convergence in iteration $t$ as  it is an iterative algorithm itself. $T^{(t-1)}_{\text{tucker}}$ depends on the convergence  \\
criteria; for R function \texttt{tucker}, the default options are \texttt{max\_iter = 25, tol = 1e-05}.\\ 
$^{\ddagger}$ 
based on Iteratively Reweighted Least Squares (IRLS) (e.g. used by R function \texttt{glm}). $T^{(t)}_{\text{glm}}$ is the\\
number of iterations to convergence and depends on convergence criteria;
for R function \texttt{glm}, the \\ 
default options are \texttt{max\_iter=25, tol=1e-08.}\\
\hline 
\end{tabular}}\vspace{-6pt}
\end{table}

\vspace{-3pt}\section{Numerical Examples}\label{sec:example}\vspace{-3pt}
We apply \our in five numerical examples (Table \ref{tab:parameter_counts}). The first three examples use simulated data on three different types of responses (Gaussian, binary, count), and the two examples are  on real-life datasets with continuous and binary responses respectively. We compare  \our with the following competing regression approaches in outcome prediction in all examples and also examine the ability of feature selection in each method in the real data examples.
\begin{itemize}
\item GLM on vectorized predictors vec($\mathcal{X}$), unregularized and with  $\ell_1$ regularization.
\item TR with CP decomposition on  $\mathcal{B}$, unregularized and with  $\ell_2$ regularization on the entries $\mathbf{U}_d$ for $d=1,2,...,D$  of  $\mathcal{B}$ after decomposed (Eq \eqref{eq:CP}) \citep{zhou2013tensor}.
\item TR with Tucker decomposition on  $\mathcal{B}$, unregularized and  with sparsity (e.g.  $\ell_1$) regularization on the core tensor \citep{li2018tucker}.  
\item Bayesian GLM on vectorized predictors vec($\mathcal{X}$) with the spike-and-slab prior on model coefficients in the simulation studies.
\item SVM in logistic TR in the simulation studies.
\item \our\hspace{-3pt} (our method): TR with Tucker decomposition on  
 $\mathcal{B}$ and $\ell_0$-regularization on the core tensor.
\end{itemize}
To make the TR models based on Tucker decomposition and CP decomposition comparable, we set the numbers of free parameters roughly the same in these two types of TR models. We first set the dimensionality of the core tensor in the Tucker decomposition and calculated the total number of parameters according to Table \ref{tab:para}. We then back-calculated the dimensionality $R$ in the CP decomposition that yields the same or  a similar number of parameters as with Tucker decomposition. The numbers of parameters in each model are listed in Table \ref{tab:parameter_counts} for the 4 examples. Though the number of parameters in the Tucker and CP decomposition-based TR in the 3 simulation studies are the same as the GLM on vec($\mathcal{X}$), the former two explore and leverage the structural information in tensor data and are expected to provide better prediction results.
\begin{table}[!htb]
\centering\vspace{-6pt}
\caption{Number parameters of different methods in the examples } \label{tab:parameter_counts}\vspace{-6pt}
\resizebox{0.78\textwidth}{!}{
\begin{tabular}{lcccc}
\hline
study & Tucker  &  CP & GLM on vec($\mathcal{X}$)\\
\hline
3 simulation studies & 64 & 60 & 64\\
&  $(R_1=R_2=R_3=4)$ & $(R=6)$ & $(I_1=I_2=I_3=4)$\\
FG-NET& 825 & 825 & 841\\
&  $(R_1=R_2=25)$ & $(R=25)$ & $(I_1=I_2=29)$\\
KTH & 64 & 60 & 64\\
&  $(R_1=R_2=R_3=4)$ & $(R=6)$ & $(I_1=I_2=I_3=4)$\\
\hline
\end{tabular}}
\resizebox{0.78\textwidth}{!}{
\begin{tabular}{l}
The dimensions $R_1$ and $R_2$ in the Tucker decomposition in the FG-NET applications\\
 were chosen by cross-validation.\\
\hline
\end{tabular}}\vspace{-9pt}
\end{table}

\vspace{-6pt}\subsection{Simulation Studies} \label{sec:simData}\vspace{-3pt}
We examine three response types $Y$ in the simulation studies -- Gaussian, binary, and count; and run linear, logistic, and Poisson TR, respectively, each case given a tensor predictor $\mathcal{X}$. Respond data $\mathbf{y}$ were generated from $\mathcal{N}(\langle\mathcal{X}, \mathcal{B}\rangle,0.5^2)$  and data on $\mathcal{X}$ were generated from $\mathcal{N}(0,1)$ in the linear TR case, $Y\sim\text{Bern}((1+e^{-\langle\mathcal{B},\mathcal{X}\rangle})^{-1})$ and data on  $\mathcal{X}$ were generated from  $\mathcal{N}(0,0.25^2)$ in the logistic TR case, and $Y\sim\text{Poisson}(e^{\langle\mathcal{B},\mathcal{X}\rangle})$ and $\mathcal{X}$ were generated from  $-2|\mathcal{N}(0.1,0.3^2)|+0.6$ in the Poisson case. Both $\mathcal{B}$ and $\mathcal{X}$ are  of dimension $4\times 4\times 4$. Elements in  $\mathcal{B}$ were randomly generated  from a certain distribution and were then adjusted to result in a sparse core tensor in the Tucker decomposition. Specifically, in the linear and logistic cases,  8 elements  of $\mathcal{B}$ were first generated from $\mathcal{N}(0,1)$, then were copied 8 times to formulate $\mathcal{B}$. The set-up of $\mathcal{B}$ in the Poisson case was similar with the only difference being that the 8 elements of $\mathcal{B}$ were generated from unif$(0,0.3)$. In all regression cases,  this way of constructing $\mathcal{B}$ leads to 2 non-zero elements and 62 zero elements  in the core tensor after the Tucker decomposition.  The sample size $n$ of each training dataset was set at $n=300$ for the linear case, $n=300$ for the binary case, and $n=200$ for the Poisson case. The size of each testing dataset was set to $200$. 

For \our\hspace{-4pt}, we set $n_e=62$, $\lambda=50$, $T=30000$, $m=600$, $\tau_0=10^{-6}$, $\tau=0.01$ in the linear TR; $n_e=62$, $\lambda=50$, $T=5000$,  $m=600$, $\tau_0=10^{-6}$, $\tau=0.01$ in the logistic TR; and $n_e=62$, $\lambda=50$, $T=10000$,  $m=600$, $\tau_0=10^{-6}$, $\tau=0.01$ in the Poisson TR.  

The TR models based on Tucker decomposition \citep{li2018tucker} use an iterative block relaxation algorithm to estimate the parameters (Algorithm \ref{al2}). Estimation of each ``block'' of parameters given the rest parameters in each iteration is equivalent to running a GLM. The objective function $l$ on lines 5 and 7 is the log-likelihood.  Two different TR models were run -- with and without $\ell_1$ regularization when estimating the core tensor. R package \texttt{glmnet} was used to run  Algorithm \ref{al2}. The hyperparameter for the $\ell_1$ regularization was set at $10^{-3}$ in the Poisson TR model (\texttt{cv.glmnet} resulted in errors)  and was chosen by 5-fold cross-validation (CV) in the other cases.

The TR models based on the  CP decomposition \citep{zhou2013tensor} used the same iterative block relaxation algorithm to estimate the parameters (Algorithm \ref{al3}).  Two different TR models were run -- with and without the $\ell_2$ regularization when estimating $\mathbf{U}_d$ for $d=1,2,...,D$ in Eq \eqref{eq:CP}.  The hyperparameter for the $\ell_2$ regularization was chosen by 5-fold CV.

For the Bayesian GLM with the spike-and-slab prior, we used R package \texttt{BoomSpikeSlab} and performed a 5-fold CV to select the optimal values for the expected model size (number non-zero elements in $\mathcal{B}$) out of $\{5, 10, 15\}$(\texttt{expected.model.size}) and the weight assigned to the prior estimates of the model parameters (\texttt{prior.information.weight}) out of $\{0.1, 0.5, 0.9\}$.

\small
\begin{algorithm}[H]
\caption{TR based on Tucker decomposition \citep{li2018tucker}}\label{al2}
\SetAlgoLined
\SetKwInOut{Input}{input}
\SetKwInOut{Output}{output}
\Input{data $(\mathcal{X}_i,y_i)$ for $i=1,\ldots,n$;iterations $T$;  stopping threshold $\eta$}
\Output{estimated  coefficient tensor $\hat{\mathcal{B}}$}
Initialize $\hat{\mathcal{G}}^{(0)}\in\mathbb{R}^{R_1\times R_2\times\dots\times R_D}$ and $\hat{\mathbf{U}}_d^{(0)}\in\mathbb{R}^{I_d\times R_d}$ for $d=1,2,...,D$\;
 $t\leftarrow 1$, convergence status $s\leftarrow 0$\;
\While{$t\leqslant T$ and $s=0$}
{
  \For{$d=1$ to $D$}
  {
$\hat{\mathbf{U}}_d^{(t)}=\argmax_{\mathbf{U}_d}\sum_{i=1}^{n}l(\hat{\mathcal{G}}^{(t-1)},\hat{\mathbf{U}}_1^{(t)},...,\hat{\mathbf{U}}_{d-1}^{(t)},{\mathbf{U}}_{d},\hat{\mathbf{U}}_{d+1}^{(t-1)},...,\hat{\mathbf{U}}_D^{(t-1)}|y_i,\mathcal{X}_i)$\;
\tcp{regularization when estimating of $\mathbf{U}_d$ can be used.}
  }
$\hat{\mathcal{G}}^{(t)}=\argmax_{\mathcal{G}}\sum_{i=1}^{n}l(\mathcal{G},\hat{\mathbf{U}}_1^{(t)},\hat{\mathbf{U}}_{2}^{(t)},...,\hat{\mathbf{U}}_D^{(t)}|y_i,\mathcal{X}_i)$\;
\tcp{regularization when estimating $\mathcal{G}$ can be used.}
$\hat{\mathcal{B}}^{(t)}=\hat{\mathcal{G}}^{(t)}\times_1\hat{\mathbf{U}}_1^{(t)}\times_2\hat{\mathbf{U}}_2^{(t)}\times_3\dots\times_\mathrm{D}\hat{\mathbf{U}}_D^{(t)}$\;
\If{$||\hat{\mathcal{B}}^{(t)}-\hat{\mathcal{B}}^{(t-1)}||_1\leqslant\eta$}{$s=1$\;}
  $t\leftarrow t+1$\;
}
Output $\hat{\mathcal{B}}^{(t)}$.\\
\end{algorithm}

\begin{algorithm}[H]
\caption{TR based on CP decomposition \citep{zhou2013tensor} }\label{al3}
\SetAlgoLined
\SetKwInOut{Input}{input}
\SetKwInOut{Output}{output}
\Input{data $(\mathcal{X}_i,y_i)$ for $i=1,\ldots,n$; maximum iterations $T$;  stopping threshold $\eta$}
\Output{estimated coefficient tensor $\hat{\mathcal{B}}$}
Initialize  $\hat{\mathbf{U}}_d^{(0)}\in\mathbb{R}^{I_d\times R}$ for $d=1,2,...,D$\;
 $t\leftarrow 1$, convergence status $s\leftarrow 0$\;
\While{$t\leqslant T$ and $s=0$}
{
  $d\leftarrow 1$\;
  \For{$d=1$ to $D$}
  {
  $\hat{\mathbf{U}}_d^{(t)}=\argmax_{\mathbf{U}_d}\sum_{i=1}^{n}l(\hat{\mathbf{U}}_1^{(t)},...,\hat{\mathbf{U}}_{d-1}^{(t)},{\mathbf{U}}_{d},\hat{\mathbf{U}}_{d+1}^{(t-1)},...,\hat{\mathbf{U}}_D^{(t-1)}|y_i,\mathcal{X}_i)$\;
  \tcp{regularization when estimating $\mathbf{U}_d$ can be used.}
  }
$\hat{\mathcal{B}}^{(t)}=\textstyle\sum_{r=1}^{R}\hat{\mathbf{u}}_1^{(t),r}\circ\hat{\mathbf{u}}_2^{(t),r}\circ\dots\circ\hat{\mathbf{u}}_D^{(t),r}$ where $\hat{\mathbf{u}}_d^{(t),r}$ is $r^{\text{th}}$ column of $\hat{\mathbf{U}}_d^{(t)}$\;
\If{$||\hat{\mathcal{B}}^{(t)}-\hat{\mathcal{B}}^{(t-1)}||_1\leqslant\eta$}{$s=1$\;}
  $t\leftarrow t+1$\;
}
Output $\hat{\mathcal{B}}^{(t)}$.\\
\end{algorithm}

\normalsize
\setstretch{1}
We run 200 repeats in each response type case and evaluated the prediction accuracy of the estimated TR models in the testing data, the accuracy of estimated or reconstructed  $\hat{\mathcal{B}}$, and the accuracy in identifying zero and non-zero elements in the core tensor after the Tucker decomposition of $\hat{\mathcal{B}}$.  
When identifying zero and non-zero elements in the core tensor for TR models not based on Tucker decomposition, Tucker decomposition was applied to obtain a core tensor in each case. We then set $\tau_0$ (the threshold for zero elements in the estimate core tensor) at 0.005, 0.05, and 0.05 for the linear, logistic, and Poisson TR, respectively. 

The results are summarized in Table \ref{tab:simulation}. In summary, \our has the smallest mean prediction error and  MSE of $\mathcal{B}$ compared to the other regression methods in all three data types. In terms of the sparsity of the core tensor, \our is the only method that can effectively promote sparsity in the core tensor. While Tucker-decomposition-based TR with the $\ell_1$ regularized core tensor also promotes core tensor sparsity, it is far from enough given there are 62 0's in the core tensor in each regression case; in addition, it runs into numerical issues during the iterative optimization (during the iterations of the block relaxation algorithm in Algorithm \ref{al2}, parameters are updated ``blockwise'' sequentially in the GLM framework with predictors vec($\mathcal{X}$) rescaled by the rest of the parameters. If a core tensor is $\ell_1$ regularized, some of the columns in the rescaled vec($\mathcal{X}$) contain only 0's, resulting in a design matrix not of full rank with ``redundant'' features). The Bayesian regression with the spike-and-slab prior promotes sparsity in a manner conceptually similar to $\ell_0$ regularization but uses probabilistic modeling to achieve sparsity without a hard penalty directly on the number of non-zero coefficients. In addition, similar to the $l_1$ regression methods in the comparison set examined in the simulation studies, the sparsity constraint is imposed on $\mathcal{B}$, which may not translate to sparsity on $\mathcal{G}$.
\begin{table}[!htb]
\centering
\caption{Performance comparison among TR models in 3 simulation studies (200 repeats) } \label{tab:simulation}\vspace{-6pt}
\resizebox{0.8\textwidth}{!}
{\begin{tabular}{lccc}
\hline
regression method & mean prediction error$^*$ & MSE$^{\#}$ & average number \\
& (SD)  &  of $\mathcal{B}$  & of 0 in core tensor \\
\hline
\multicolumn{4}{c}{\cellcolor{gray!25}{linear TR}}\\
\hline
\our & \textbf{0.4114} (0.0181)  & \textbf{0.0002}& \textbf{62.00} \\
unregularized Tucker & 0.4487  (0.0213) & 0.0011 & 17.50 \\
$\ell_1$ regularized Tucker & 0.4393  (0.0211) & 0.0008 & 28.64 \\
unregularized CP & 0.4415 (0.0205) & 0.0009 & 23.22 \\
$\ell_2$ regularized CP & 0.4269 (0.0203) & 0.0006 & 26.06 \\
vectorized & 0.4487 (0.0213) & 0.0011 & 17.50 \\
$\ell_1$ regularized vectorized & 0.4490 (0.0214) & 0.0011 & 17.46 \\
Bayesian spike-slab & 0.4484  (0.0212) & 0.0324 & 17.02 \\
\hline
\multicolumn{4}{c}{\cellcolor{gray!25}{Poisson TR}}\\
\hline
\our  & \textbf{1.5639} (0.1110) & \textbf{0.0051} &  \textbf{62.00} \\
unregularized Tucker & 1.8391 (0.1626) & 0.0165 & 25.22 \\
$\ell_1$ regularized Tucker$^{\$}$  & 1.8272 (0.1625) & 0.0160 & 25.90 \\
unregularized CP & 1.8154 (0.1593) &  0.0156 & 27.20 \\
$\ell_2$ regularized CP & 1.6196 (0.1207) & 0.0074 & 42.52 \\
vectorized MLE & 1.8391 (0.1626) & 0.0165 & 25.23 \\
$\ell_1$ regularized MLE & 1.7307 (0.1216) & 0.0136 & 28.35 \\
Bayesian spike-slab & 2.0233  (0.1637) & 0.1793 & 22.38 \\
\hline
\multicolumn{4}{c}{\cellcolor{gray!25}{logistic TR}}\\
\hline
\our & \textbf{28.67\%} (0.0246) & \textbf{0.1055} & \textbf{62.00} \\
unregularized Tucker & 29.86\% (0.0288) & 0.8672 & 3.98  \\
$\ell_1$ regularized Tucker & 30.18\% (0.0357) & 0.6872 & 16.62 \\
unregularized CP  & 29.85\% (0.0207) & 0.7542 & 3.00 \\
$\ell_2$  regularized CP & 29.18\% (0.0259) & 0.2378 & 13.26 \\
vectorized & 29.86\% (0.0288) & 0.8672 & 3.98 \\
$\ell_1$  regularized vectorized & 33.19\% (0.0089) & 0.3014 & 7.99 \\
Bayesian spike-slab & 37.41\%  (0.0436) & 0.6908 & 11.11\\
SVM  & {33.84\%} (0.0024) &-  &- \\
\hline
\end{tabular}}
\resizebox{0.8\textwidth}{!}{\begin{tabular}{l}
\textbf{Bold numbers} represent the best performer by each metric in each TR model case.\\
$^*$ averaged $\ell_1$ prediction error $(200n)^{-1}\sum_{j=1}^{200}\sum_{i=1}^n|\hat{y}_{ij}-y_i|$ in testing data for linear \\ 
\hspace{6pt}  and  Poisson TR,  where $\hat{y}_{ij}$  is the  prediction for observation $i$ in repeat $j$; averaged \\
\hspace{6pt}  misclassification rate  over 200 repeats  in testing  data for logistic TR  ($\hat{y}_{ij}\!=\!1$ if   \\ 
\hspace{6pt} $\hat{\Pr}(y_{ij}\!=\!1)\!>\!0.5$, where $\hat{\Pr}(y_{ij}\!=\!1)$ is for  observation $i$ in repeat $j$,  $\hat{y}_{ij}\!=\!0$ otherwise.)  \\
$^{\#}$ mean squared error $\sum_{i=1}^{4}\sum_{j=1}^{4}\sum_{k=1}^{4}(\hat{b}_{ijk}-b_{ijk})^2/4^3$, where $b_{ijk}$ is the  true value of \\\hspace{6pt} element  $(i,j,k)$   in $\mathcal{B}$, and $\hat{b}_{ijk}$ is its estimate averaged over 200 repeats.\\  
$^{\$}$ The $\ell_1$ regularized Poisson TR model with Tucker decomposition had convergence \\ \hspace{6pt}  problems. The metrics were calculated using  180 out of 200 repeats that  converged. \\
\hline
\end{tabular}}\vspace{-6pt}
\end{table}

\vspace{-6pt}\subsection{Real Data Application} 
We applied  different TR methods and compared their performances in two real datasets. The first is the FG-NET dataset that contains human face images \citep{ranking2014ECCV} (\url{http://yanweifu.github.io/FG_NET_data/FGNET.zip}); the task is age prediction given the images and identification of face areas (image pixels) that are relevant for age prediction. The second application is the KTH dataset that contains human motion videos \citep{1334462} (\url{https://www.csc.kth.se/cvap/actions/}) and we aim for classification of upper vs. lower body motions and identification of video regions that are relevant for the prediction.

\our is the best in terms of both prediction accuracy and feature selection ability in the FG-NET  example and in feature selection in the KTH dataset and its performance is not the best or the worst in prediction accuracy in the  KTH dataset.  In what follows, we describe the data, the applications, and the results in detail.

\subsubsection{Facial Age Prediction} 
\label{sec:real:fgnet}
The FG-NET dataset contains 1,002 human face images (color or greyscale) from $82$ individuals aged from $0$ to $69$ years old. Each image is labeled with the person's age and contains $68$ manually annotated points, which  describe the structure of a face and are located around the eyebrows, mouth, eyes, nose, and the edge of a face. We examined the performances of different TR methods in the We run linear TR with $l_2$ loss to predict age given an image and identify pixels that are important predictors of age.

\begin{figure}[!htp]
\centering
\includegraphics[width=2.3cm, height=2.5cm]{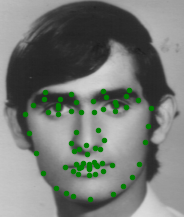}\hspace{6pt}
\includegraphics[width=2.3cm, height=2.5cm]{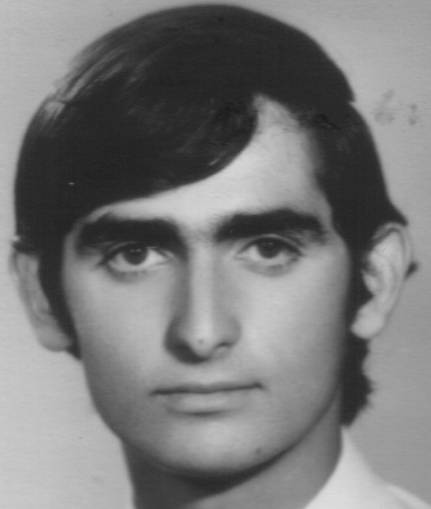}\hspace{6pt}
\includegraphics[width=2.3cm, height=2.5cm]{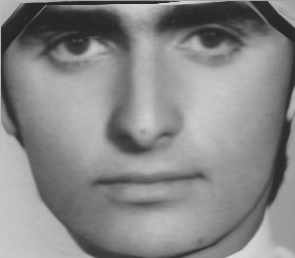}\hspace{6pt}
\includegraphics[width=2.3cm, height=2.5cm]{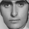}
\\[\smallskipamount]
\includegraphics[width=2.3cm, height=2.5cm]{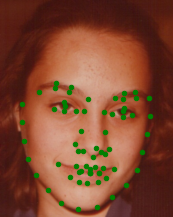}\hspace{6pt}
\includegraphics[width=2.3cm, height=2.5cm]{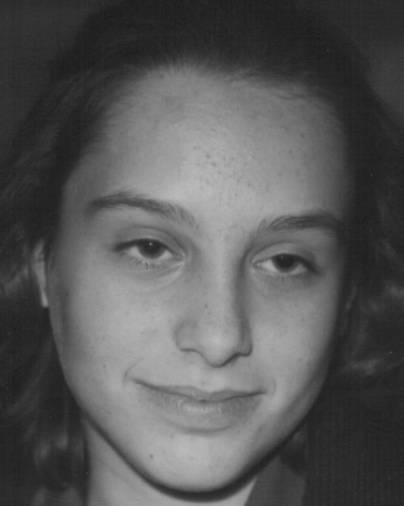}\hspace{6pt}
\includegraphics[width=2.3cm, height=2.5cm]{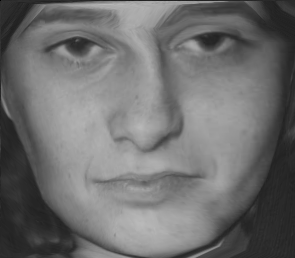}\hspace{6pt}
\includegraphics[width=2.3cm, height=2.5cm]{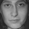}
\\[\smallskipamount]
\hspace{2pt}
\subcaptionbox{\footnotesize{original with landmarks}}{
\includegraphics[width=2.3cm, height=2.5cm]{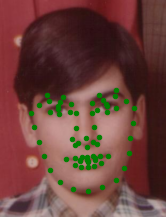}}\hspace{2pt}
\subcaptionbox{\footnotesize{grayscale}}{
\includegraphics[width=2.3cm, height=2.5cm]{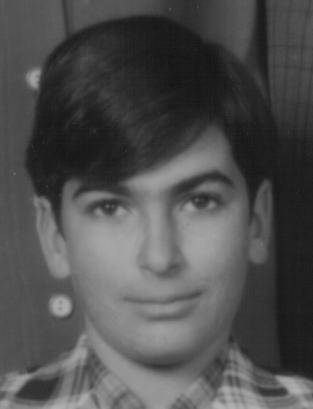}}\hspace{6pt}
\subcaptionbox{\footnotesize{after piecewise affine transformation}}{\includegraphics[width=2.3cm, height=2.5cm]{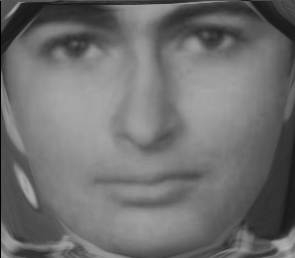}}\hspace{3pt}
\subcaptionbox{\footnotesize{rescaled}}{
\includegraphics[width=2.3cm, height=2.5cm]{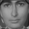}}\hspace{6pt}
\vspace{-6pt}\caption{Pre-processing of the FG-NET images} 
\label{fig:preprocessing-fg} \vspace{-9pt}
\end{figure}

We first pre-processed the images  before the model fitting. Specifically, we transformed all images to grayscale and  used annotated points for a piecewise affine transformation that aligned the points and corners across all images. For example, the tip of the nose is located in the same position in a 2D Cartesian coordinate (xy) after transformation in each image; the same for all other points. We then stacked all transformed images to form a 3D block along the z-axis. The pre-processing procedure is illustrated with 3 examples in Figure \ref{fig:preprocessing-fg}.

We trained linear TR models on the final pre-processed gray-scaled $29\times 29$ images to predict age and to identify pixels that are important for age prediction ($800$ images were used as the training set, $100$ images as the validation set, and the rest as the testing set). 

For the \our procedure, we set $n_e=615, \lambda=50$, $T=10000$, $m=600$, $\tau_0=10^{-6}$, and $\tau=0.01$. We used the validation set to select hyperparameters $n_e$ and core tensor dimensions $R_1$ and $R_2$ (Table \ref{tab:parameter_counts}) for \our (\our is relatively robust to a wide range of $\lambda$; in the two real data examples, $\lambda$ from 50 to 500 yield similar results). For other TR models that involve regularization, we merged the training and validation set and used a five-fold CV to choose the hyperparameters. For TR models without regularization, we merge the training and validation set for training.

We calculated the mean absolute prediction error $\sum_{i=1}^{n}|\hat{y}_i-y_i|/n$ in both real applications, where $n$ is the number of images in the testing data, $y_i$ is the observed response, and $\hat{y}_i$ is predicted response from each of the fitted TR models. The results are presented in Table \ref{tab:age}.  \our yields the smallest prediction error with a mean error $3.63$ years, followed by the $l_1$ regularized GLM on vectorized $\mathcal{X}$ and $\mathcal{B}$, whereas the prediction errors in other TR models, regularized or not, are almost doubled or even quintupled. 
\begin{table}[!htb]
\centering\caption{Predicted errors in different models in age prediction}\label{tab:age}\vspace{-6pt}
\resizebox{1\textwidth}{!}{
\begin{tabular}{c@{\hspace{3pt}}c@{\hspace{5pt}}c@{\hspace{6pt}}c@{\hspace{6pt}}c@{\hspace{6pt}}c@{\hspace{3pt}}c}
\hline
\our & $\ell_1$ regularized  & $\ell_1$-regularized & $\ell_2$-regularized  & unregularized & unregularized  & GLM 
\\
 & GLM  on vec($\mathcal{X}$) & Tucker &  CP &  CP & Tucker &  on vec($\mathcal{X}$)\\
\hline
\textbf{3.63} & 4.30 & 7.11 & 9.51 & 20.57 & 20.57 & 20.80 \\

\hline
\end{tabular}}\vspace{-12pt}
\end{table}

\begin{figure}[!htb]
\raggedleft
\hspace{-3pt}\subcaptionbox{\scriptsize\our}{
\includegraphics[scale=0.23]{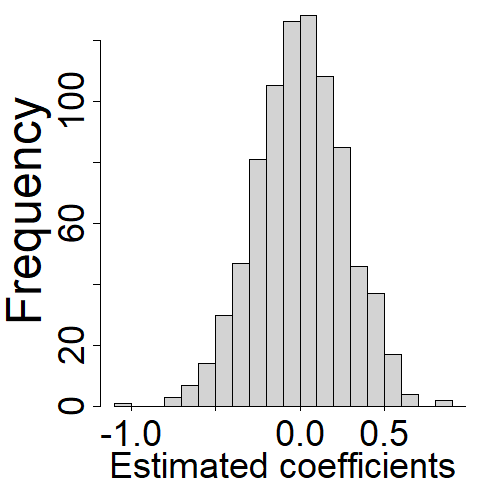}}
\subcaptionbox{\scriptsize{l$_1$ GLM}}{
\includegraphics[scale=0.23]{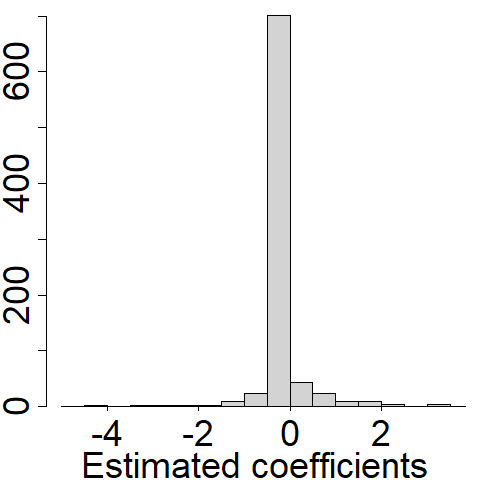}}
\subcaptionbox{\scriptsize{Tucker+lasso}}{
\includegraphics[scale=0.23]{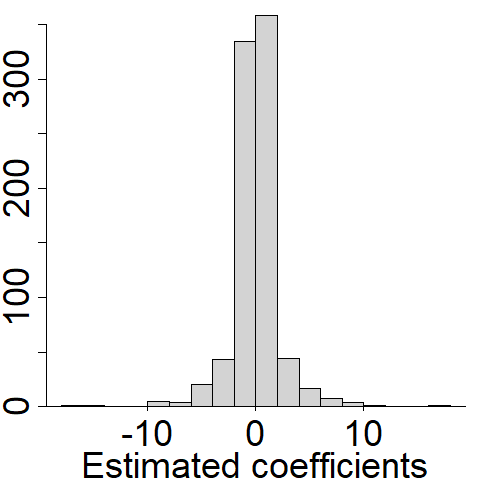}}
\subcaptionbox{\scriptsize{CP+$\ell_2$}}{ 
\includegraphics[scale=0.23]{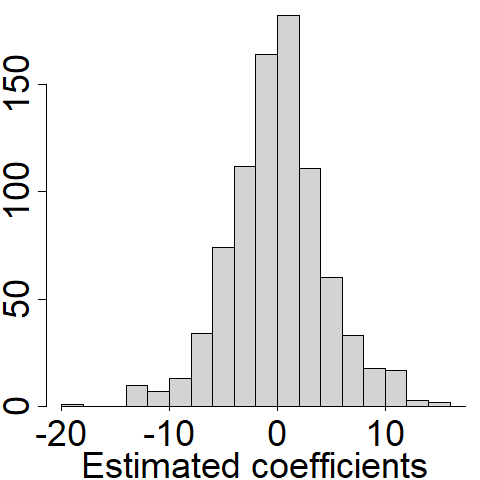}}\\
\subcaptionbox{\scriptsize{CP}}{
\includegraphics[scale=0.23]{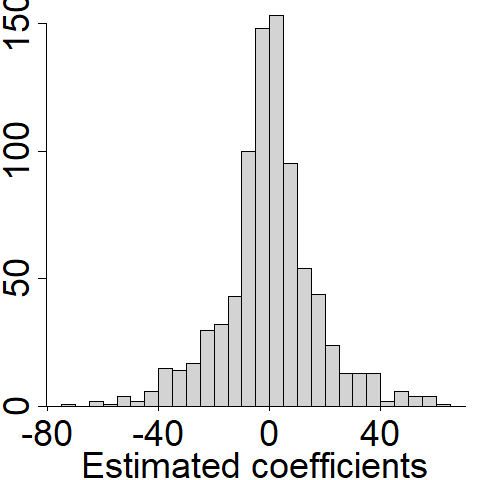}}
\subcaptionbox{\scriptsize{Tucker}}{
\includegraphics[scale=0.23]{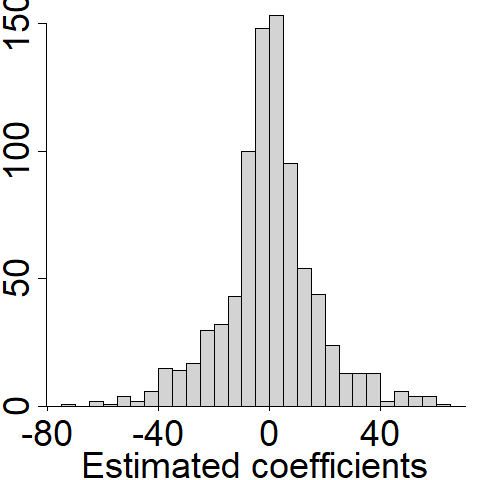}}
\subcaptionbox{\scriptsize{GLM}}{
\includegraphics[scale=0.23]{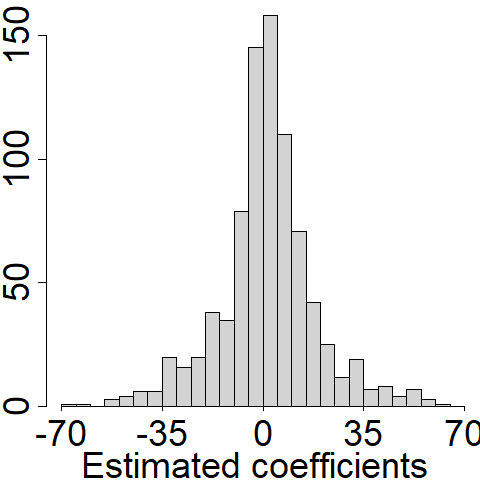}}\vspace{-3pt}
\caption{Histograms of estimated coefficients in different methods}
\label{fig:beta_hist}\vspace{-9pt}
\end{figure}

We also examined the ability of TR models on their ability to identify important pixels in age prediction based on the estimated $\hat{\mathcal{B}}$. Note that among all the TR models used, only the $\ell_1$-regularized GLM sets some of the elements in estimated $\hat{\mathcal{B}}$ at 0. Though the TR model  based on the Tucker decompositions uses $\ell_1$ regularization and \our achieves $\ell_0$ regularization, the sparsity regularizations are imposed on the core tensor after Tucker decomposition of $\mathcal{B}$ rather than $\mathcal{B}$ itself, and the sparsity of the core tensor does not necessarily translate into sparsity of  $\hat{\mathcal{B}}$, which is constructed from the sparse core tensor and factor matrices $\mathbf{U}$'s. Figure \ref{fig:beta_hist} depicts the histograms of the elements in estimated $\hat{\mathcal{B}}$ in each TR model.  The magnitude of the elements in $\hat{\mathcal{B}}$ varies significantly across different methods. The estimated elements from \our and  $\ell_1$-regularized GLM  are more similar in magnitude (ranging
from around $-1$ to 1 in the former and -4 to 3 in the latter) while the other methods can produce extremely large values in $\hat{\mathcal{B}}$ (e.g, $\pm70$).

\begin{figure}[!htb]
\small{ \our\hspace{1cm} $\ell_1$ GLM \hspace{1.3cm}   $\ell_1$ Tucker \hspace{1.7cm} $\ell_2$  CP\hspace{1.8cm} CP \hspace{1.5cm} Tucker}\\
\includegraphics[width=2cm, height=2cm]{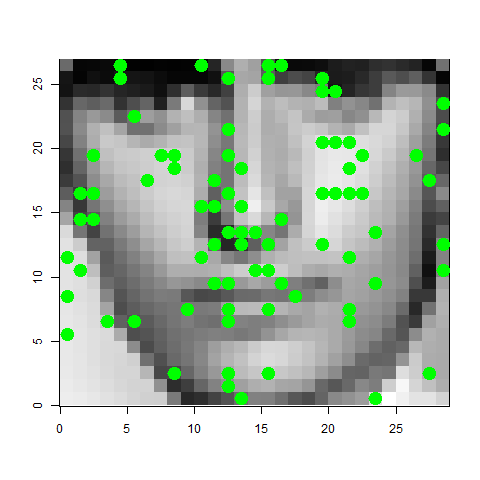}\hfill
\includegraphics[width=2cm, height=2cm]{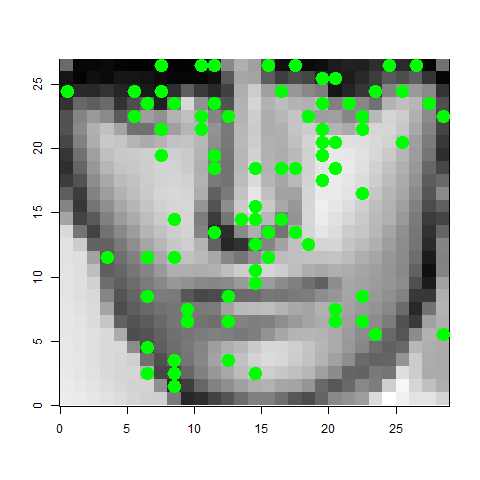}\hfill
\includegraphics[width=2cm, height=2cm]{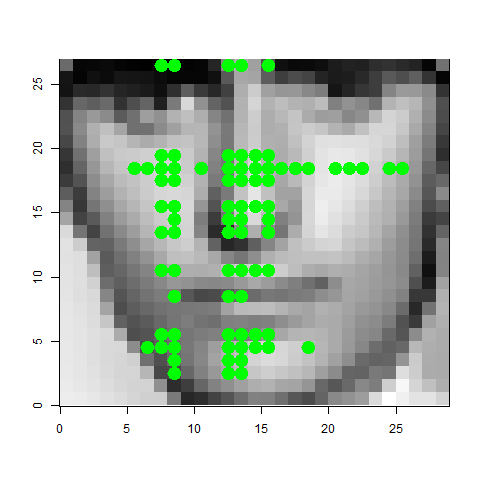}\hfill
\includegraphics[width=2cm, height=2cm]{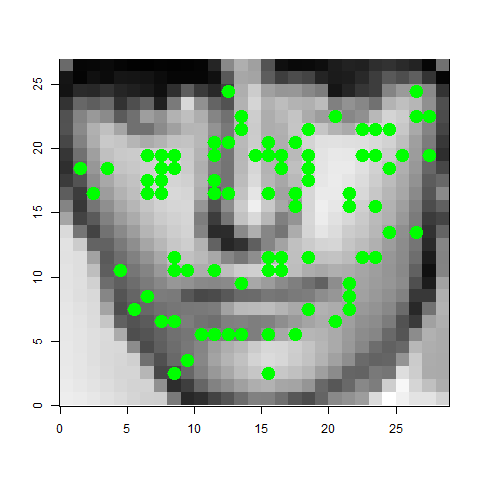}\hfill
\includegraphics[width=2cm, height=2cm]{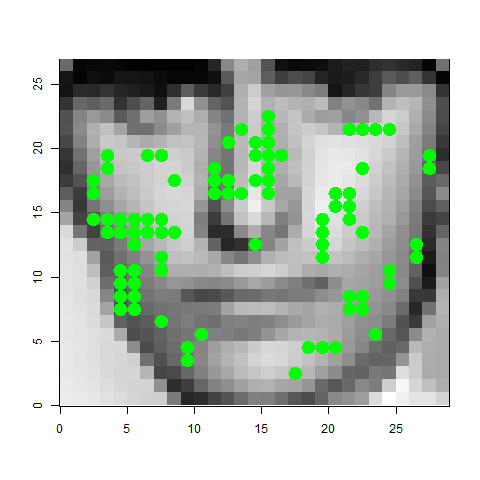}\hfill
\includegraphics[width=2cm, height=2cm]{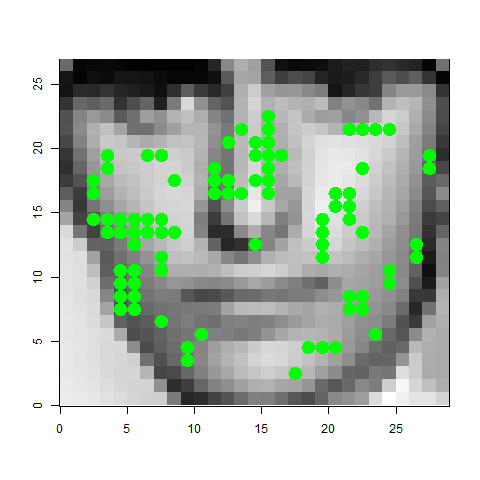}\hfill\\
\hspace*{2pt}
\includegraphics[width=2cm, height=2cm]{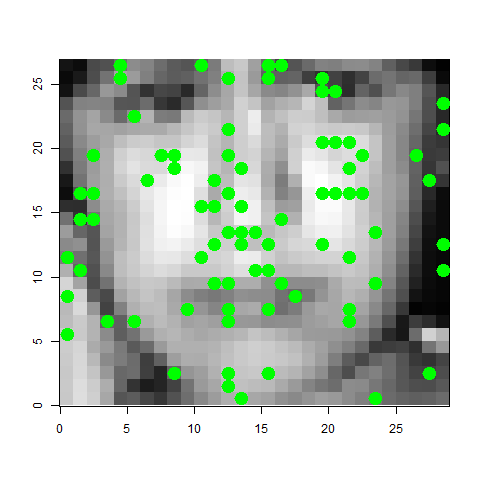}\hfill
\includegraphics[width=2cm, height=2cm]{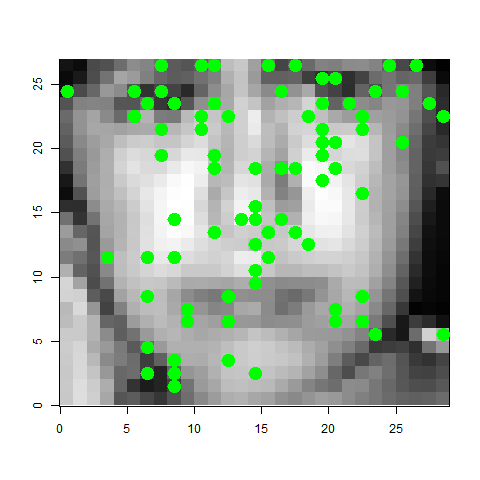}\hfill
\includegraphics[width=2cm, height=2cm]{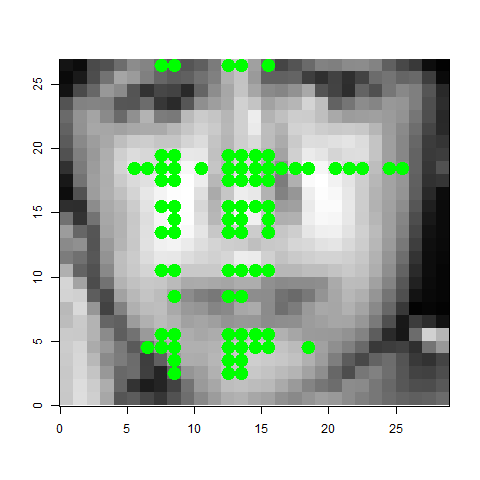}\hfill
\includegraphics[width=2cm, height=2cm]{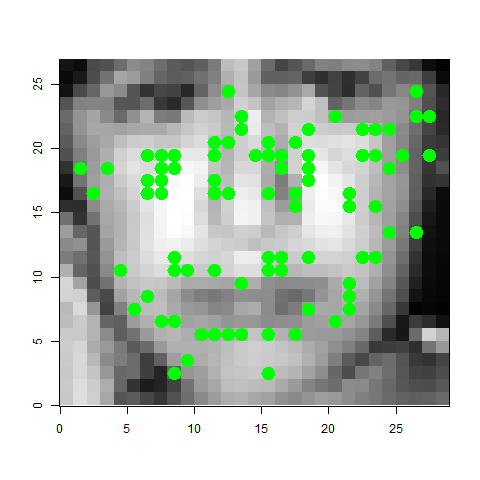}\hfill
\includegraphics[width=2cm, height=2cm]{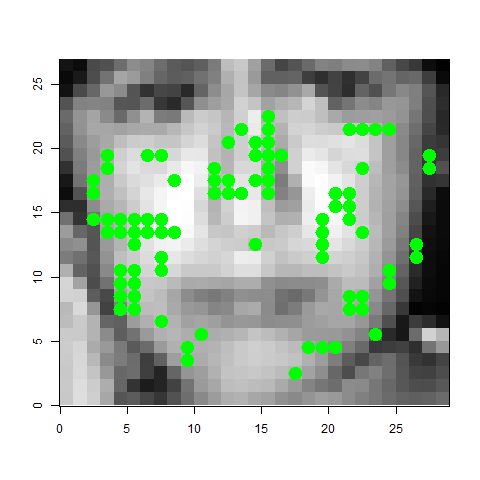}\hfill
\includegraphics[width=2cm, height=2cm]{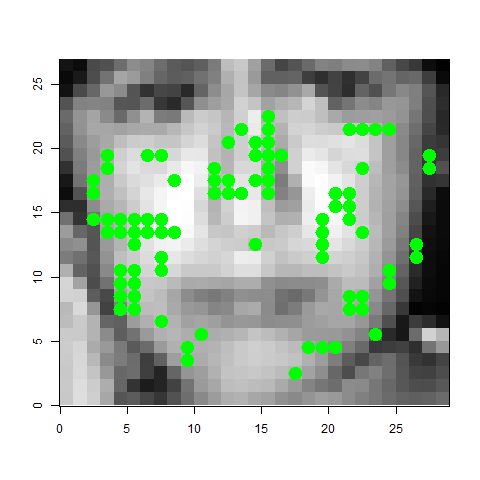}\hfill\\
\hspace*{-2pt}
\includegraphics[width=2cm, height=2cm]{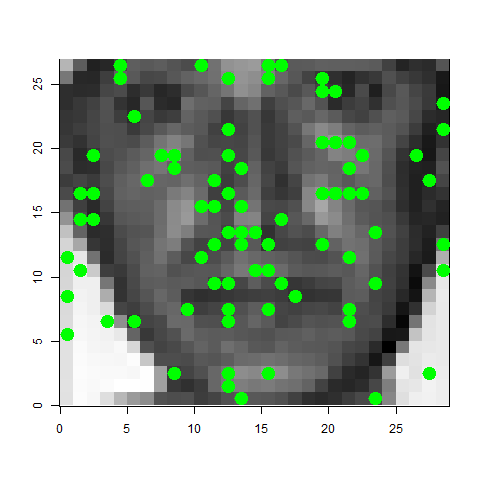}\hfill
\includegraphics[width=2cm, height=2cm]{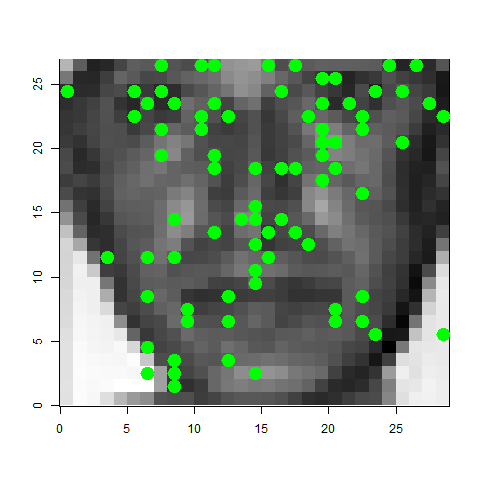}\hfill
\includegraphics[width=2cm, height=2cm]{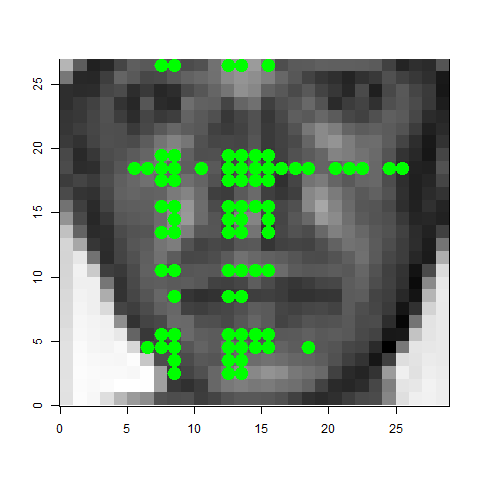}\hfill
\includegraphics[width=2cm, height=2cm]{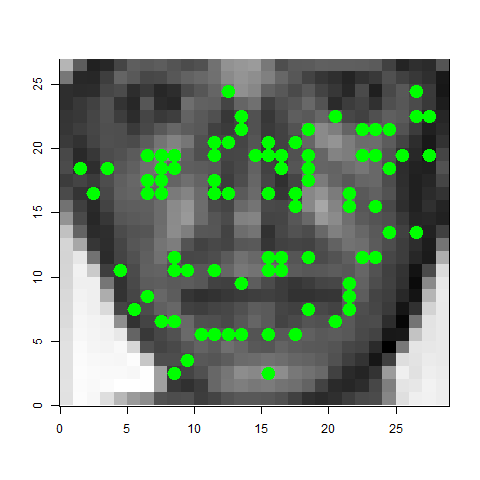}\hfill
\includegraphics[width=2cm, height=2cm]{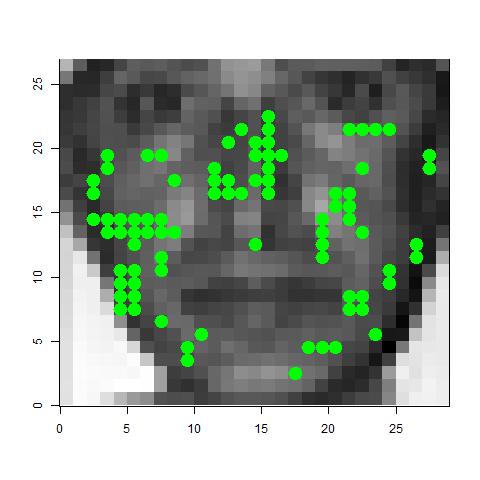}\hfill
\includegraphics[width=2cm, height=2cm]{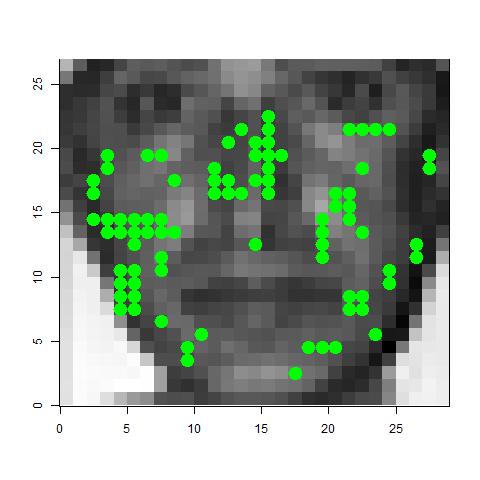}\hfill
\\
\vspace{-4pt}\centering (a) pixels corresponding to the top 10\% elements in $|\hat{\mathcal{B}}|$ are plotted\\

\hspace{2pt}
\includegraphics[width=2cm, height=2cm]{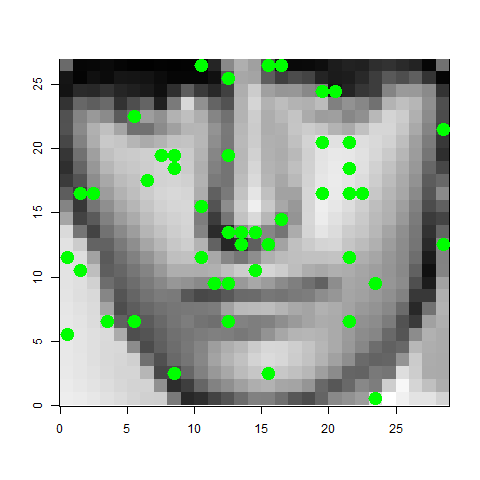}\hfill
\includegraphics[width=2cm, height=2cm]{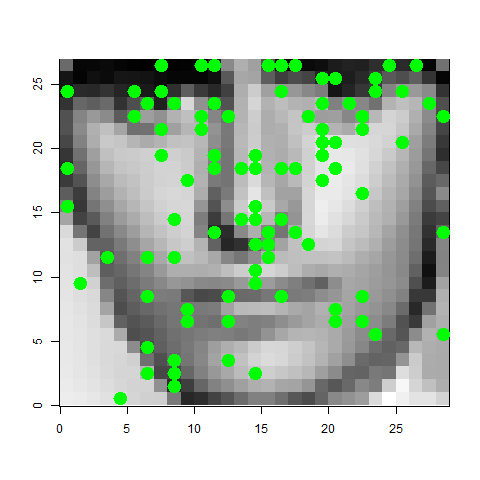}\hfill
\includegraphics[width=2cm, height=2cm]{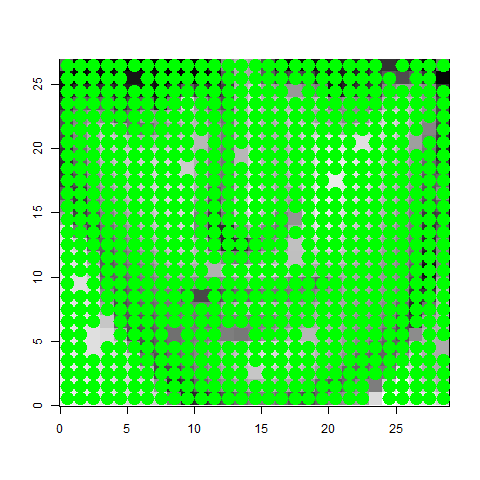}\hfill
\includegraphics[width=2cm, height=2cm]{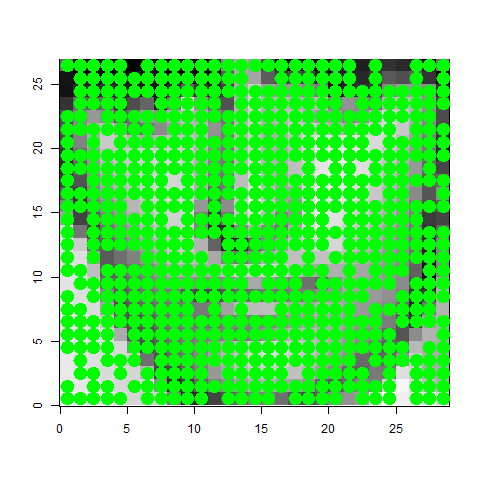}\hfill
\includegraphics[width=2cm, height=2cm]{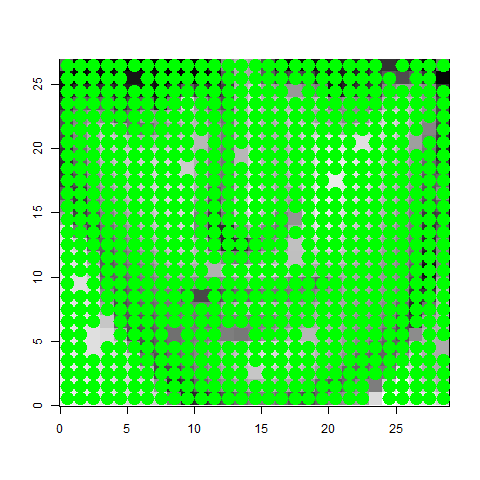}\hfill
\includegraphics[width=2cm, height=2cm]{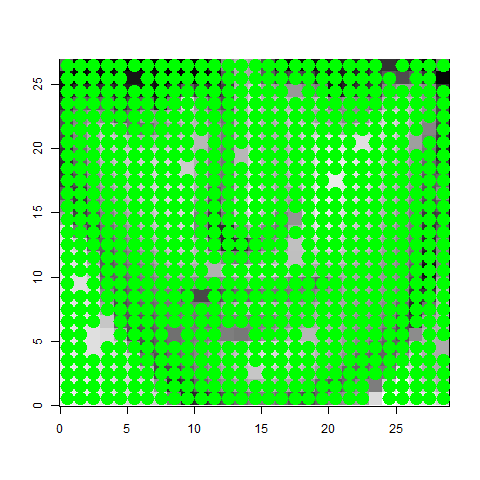}\hfill
\\
\hspace*{2pt}
\includegraphics[width=2cm, height=2cm]{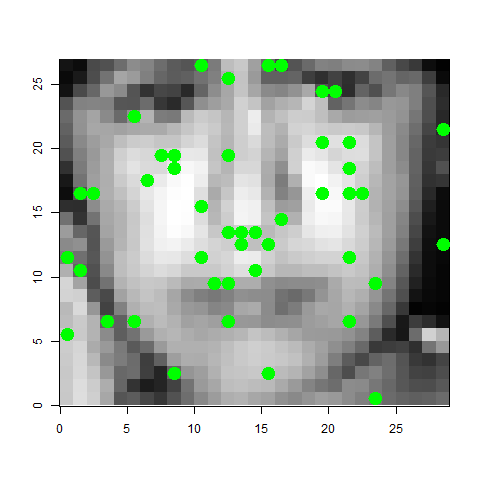}\hfill
\includegraphics[width=2cm, height=2cm]{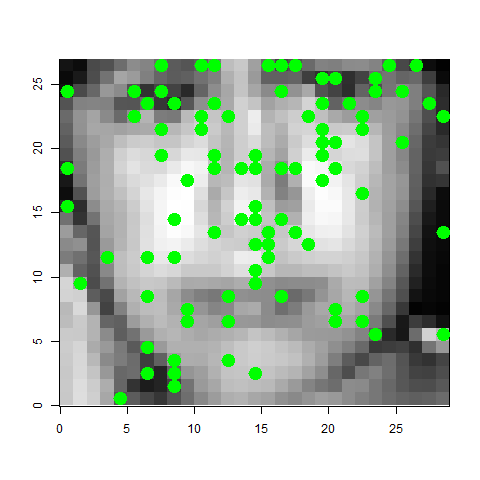}\hfill
\includegraphics[width=2cm, height=2cm]{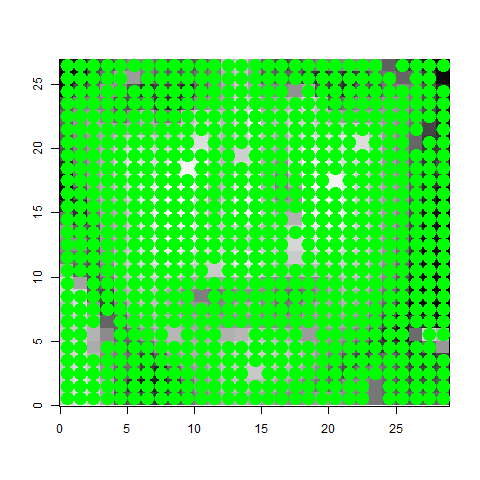}\hfill
\includegraphics[width=2cm, height=2cm]{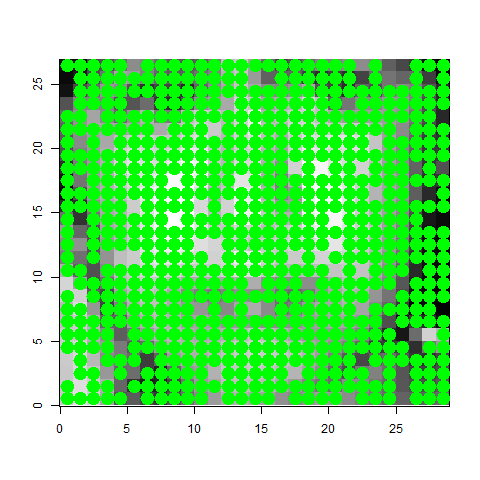}\hfill
\includegraphics[width=2cm, height=2cm]{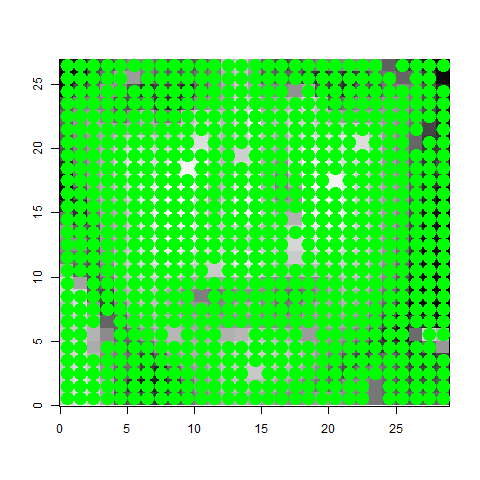}\hfill
\includegraphics[width=2cm, height=2cm]{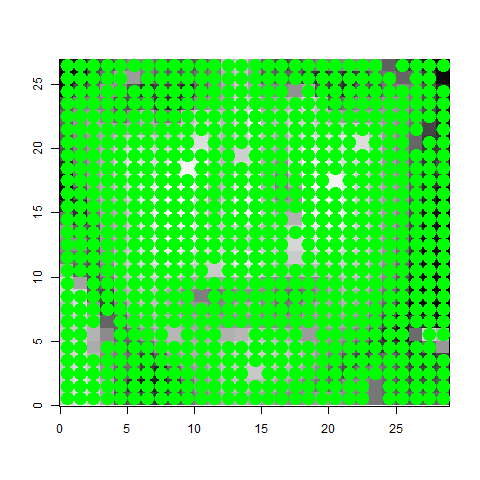}\hfill\\
\hspace*{-2pt}
\includegraphics[width=2cm, height=2cm]{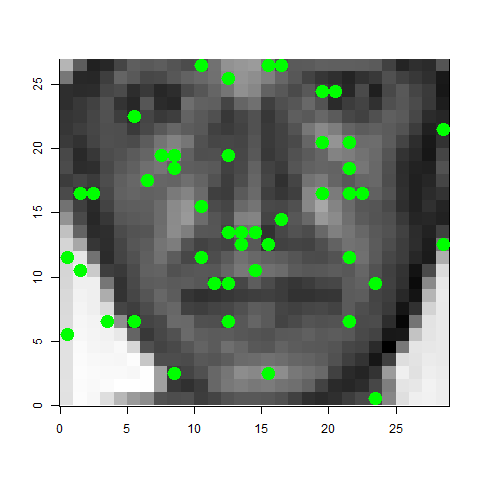}\hfill
\includegraphics[width=2cm, height=2cm]{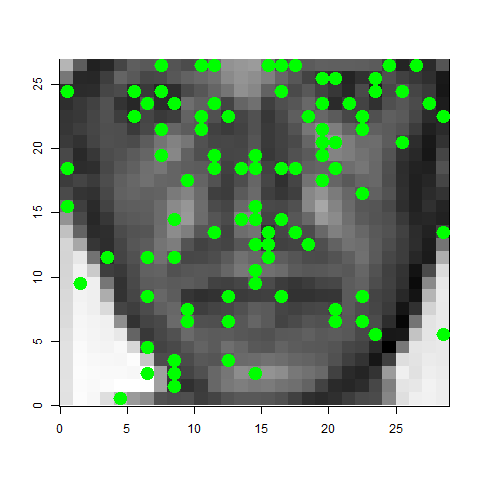}\hfill
\includegraphics[width=2cm, height=2cm]{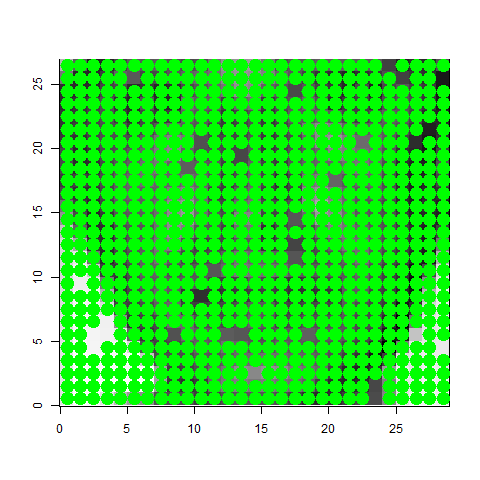}\hfill
\includegraphics[width=2cm, height=2cm]{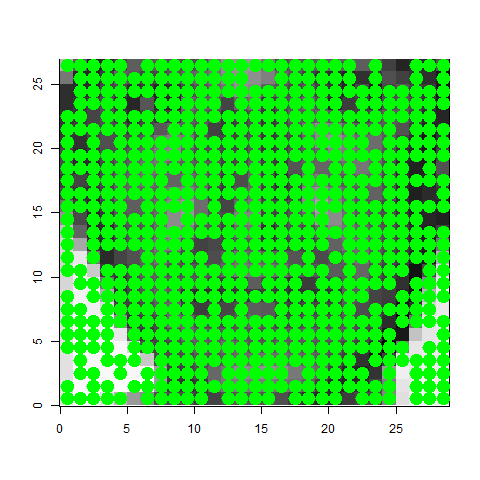}\hfill
\includegraphics[width=2cm, height=2cm]{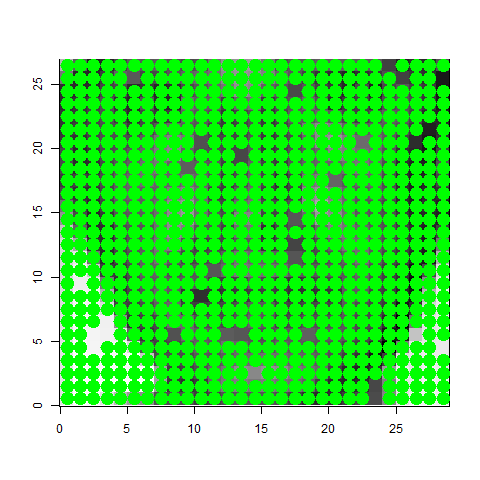}\hfill
\includegraphics[width=2cm, height=2cm]{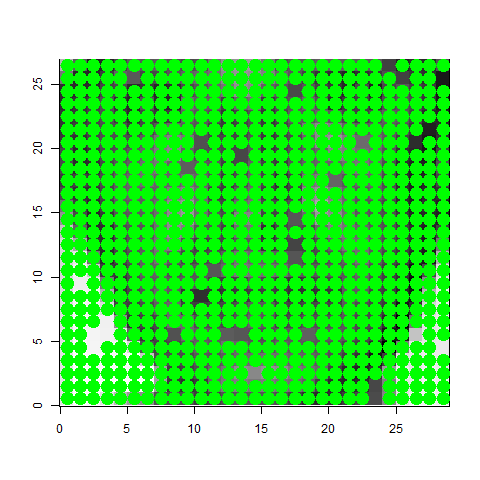}\hfill

\vspace{-3pt}
\centering\normalsize (b) pixels corresponding to  elements in $|\hat{\mathcal{B}}| > 0.5$ are plotted\vspace{-3pt}
\caption{Relevant pixels (green dots)  for age prediction by different methods identified by two hard thresholding methods}
\label{fig:fg}\vspace{-12pt}
\end{figure}

We took the top 10\%  elements in $\hat{\mathcal{B}}$ in absolute value in each of the regression methods  and plotted the corresponding pixels in Figure \ref{fig:fg}(a). We also applied hard thresholding at 0.5 to the absolute values of the elements in $\hat{\mathcal{B}}$ and plotted  the  pixels with corresponding elements in $|\hat{\mathcal{B}}|>0.5$ in Figure \ref{fig:fg}(b). In both cases, the sparsity regularization imposed by \our  on the core tensor, though not on $\mathbf{B}$ directly, helps to locate pixels  important for age prediction,  which are concentrated around eyebrows, eyes,  the area under the eyes, cheekbones, laugh lines, and the area around the mouth, and jawlines, which are all related to the early signs of aging (\url{https://www.skinmd.ph/blog-content/2017/12/8/the-anatomy-of-an-aging-face}). The $l_1$ regularized linear regression with vectorized predictors also identifies similar important pixels for age predictions, many of which overlap those identified  by \our but are more sparse. 
By contrast, the other TR models, regularized or not,  select pixels that are not quite meaningful in terms of age prediction in Figure \ref{fig:fg}(a), or select overwhelmingly too many pixels in the hard-thresholding approach and do not help identify the important signs for age prediction.

\subsubsection{Human Action Recognition} 
\label{sec:real:kth}
The KTH dataset  contains black-white videos of 25 human subjects doing 6 different types of actions (walking, jogging, running, boxing, hand waving and hand clapping) in 4 different scenes, leading to a total of $25\times 6\times 4=600$ videos. Each video records a sequence of the same subject performing the same action in the same scene that lasts around 5 seconds and has 25 fps frame rate. Each sequence consists of a set of frames, and each frame can be viewed as a single image. Each sequences was used as one training/testing sample in our experiment. After removing sequences that don't have correct starting/ending frame stamps in a video, we ended with a total of 2259 sequences, which was split to 1697 training samples from 19 subjects and 562 testing samples from the remaining 6 subjects.  For each sequence, we sampled 4 frames that were evenly distributed along the time dimension (first and last frames were always included), and rescaled each frame from the original resolution of $160\times 120$ pixels to $4\times 4$. We divided each pixel value by 255 to normalize it to $[0,1]$. 

We grouped the videos types into upper body motion (1115 sequences including boxing, hand waving and hand clapping) and lower body motion (1144 sequences including walking, jogging, running), and run logistic TR on the $4\times 4\times 4$ videos  with negative likelihood loss to predict 2-class motion type. For the \our procedure, we set $n_e=32, \lambda=50$, $T=3000$, $m=600$, $\tau_0=10^{-6}$, and $\tau=0.01$. The core tensor in \our and Tucker models (both regularized and unregularized) were of dimension $4\times 4\times 4$. For CP models (both regularized and unregularized), $R=6$. The reason for choosing such dimensions is explained in Section \ref{sec:example} and Table \ref{tab:parameter_counts}.

\begin{figure}[!htb]
\footnotesize{Frame \hspace{0.1cm} $\ell_2$  CP\hspace{0.8cm} $\ell_1$  GLM  \hspace{0.8cm} EN Tucker \hspace{0.4cm} \our \hspace{0.6cm} Tucker \hspace{1cm} CP}\\
\centering
\raisebox{0.9cm}{1. }{
\includegraphics[width=2cm, height=2cm]{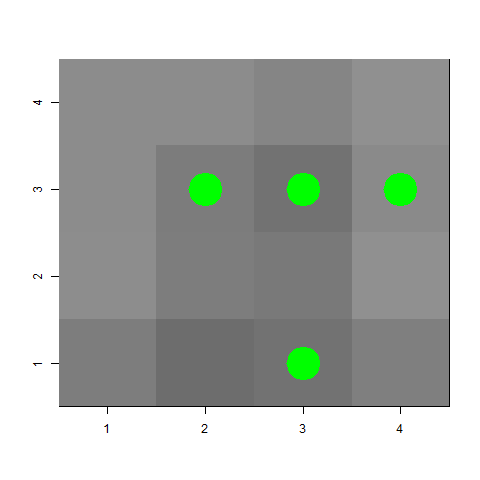}
\includegraphics[width=2cm, height=2cm]{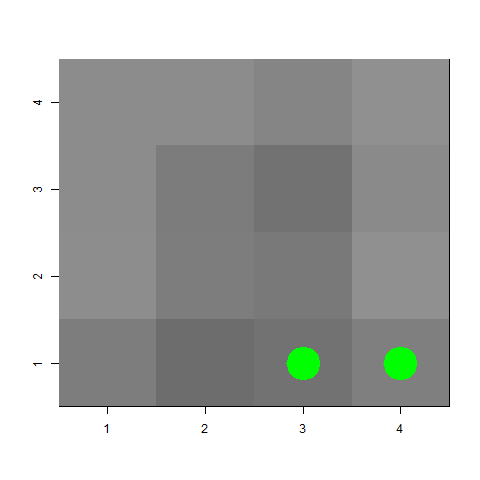}
\includegraphics[width=2cm, height=2cm]{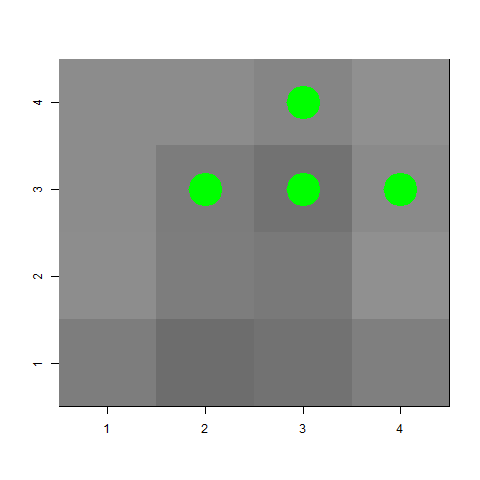}
\includegraphics[width=2cm, height=2cm]{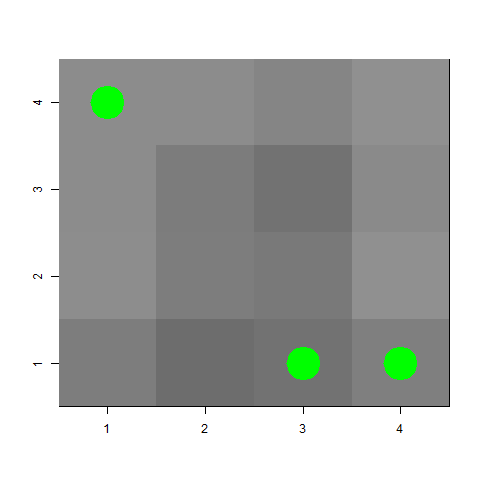}
\includegraphics[width=2cm, height=2cm]{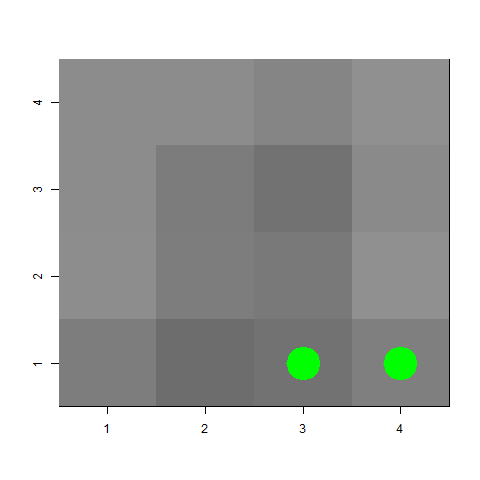}
\includegraphics[width=2cm, height=2cm]{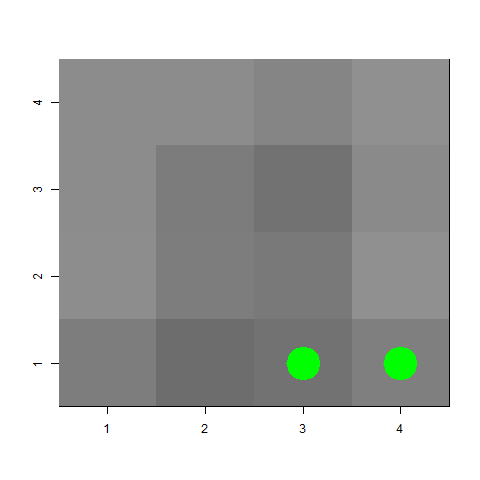}}
\\
\hspace*{2pt}
\raisebox{0.9cm}{2. }{
\includegraphics[width=2cm, height=2cm]{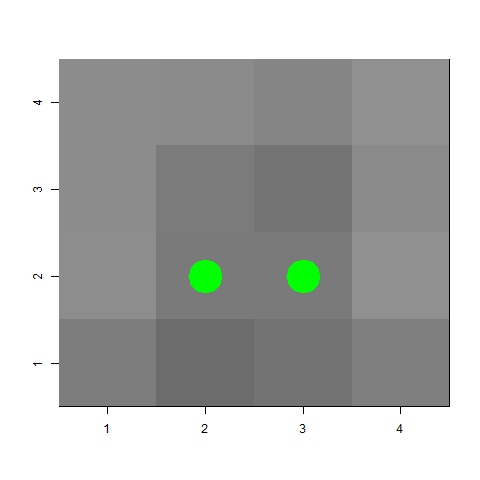}
\includegraphics[width=2cm, height=2cm]{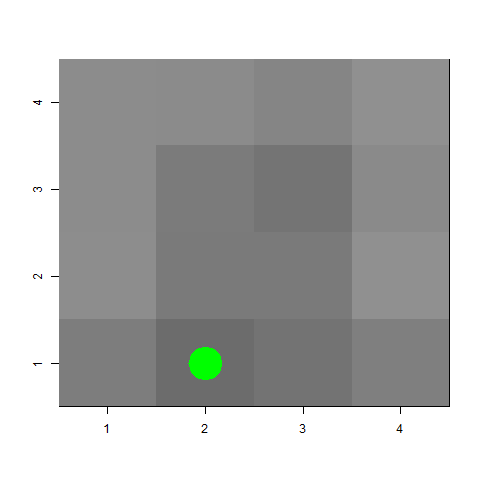}
\includegraphics[width=2cm, height=2cm]{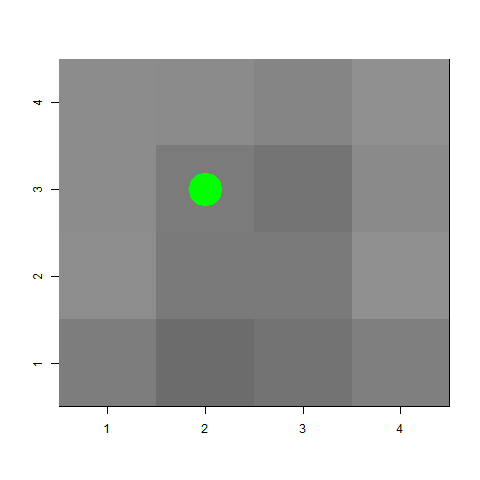}
\includegraphics[width=2cm, height=2cm]{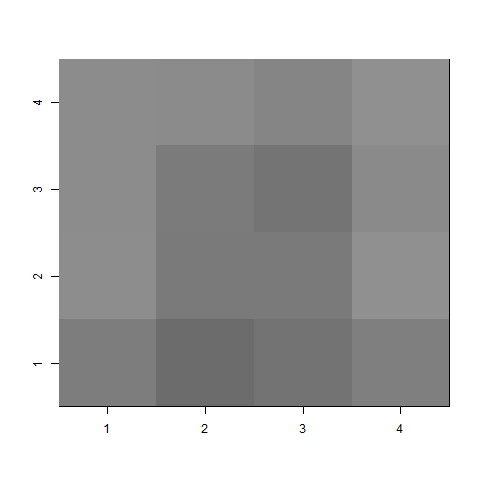}
\includegraphics[width=2cm, height=2cm]{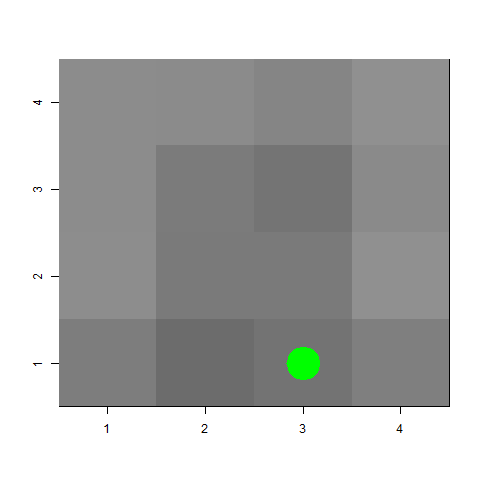}
\includegraphics[width=2cm, height=2cm]{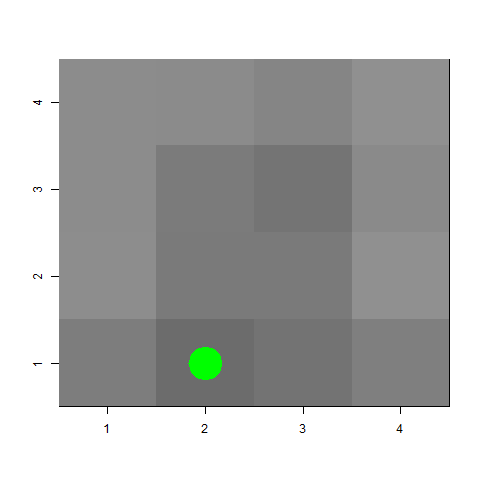}}
\\
\hspace*{2pt}
\raisebox{0.9cm}{3. }{
\includegraphics[width=2cm, height=2cm]{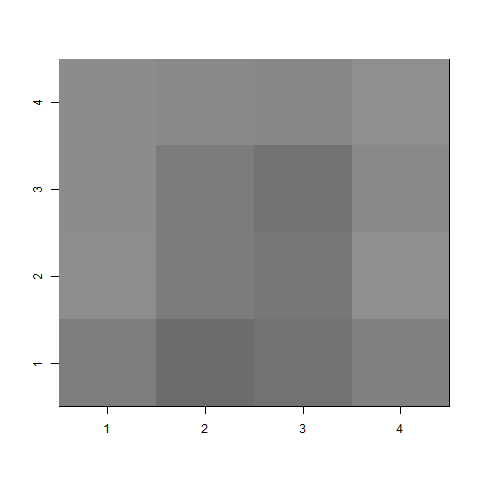}
\includegraphics[width=2cm, height=2cm]{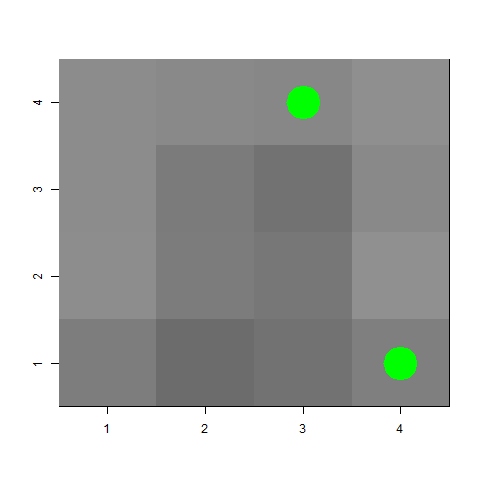}
\includegraphics[width=2cm, height=2cm]{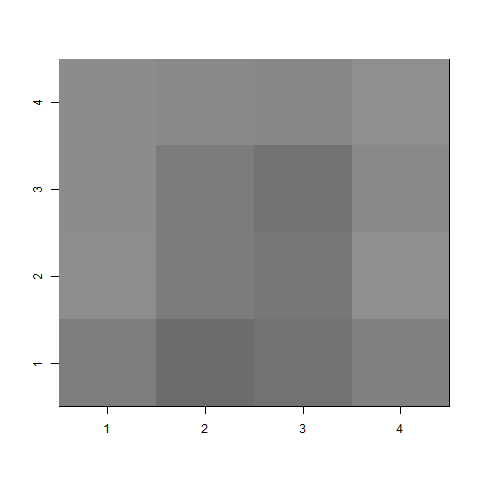}
\includegraphics[width=2cm, height=2cm]{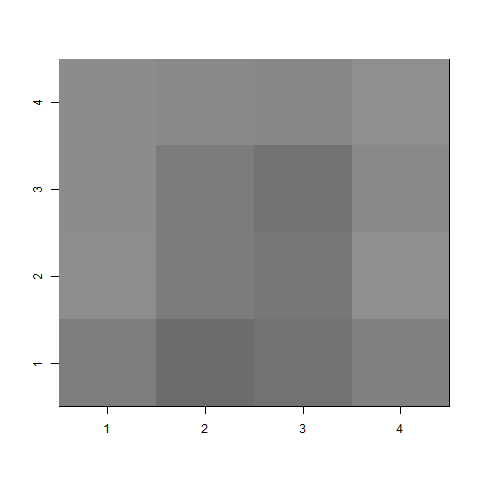}
\includegraphics[width=2cm, height=2cm]{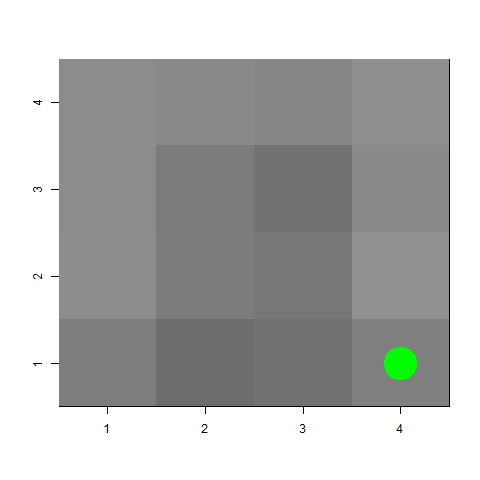}
\includegraphics[width=2cm, height=2cm]{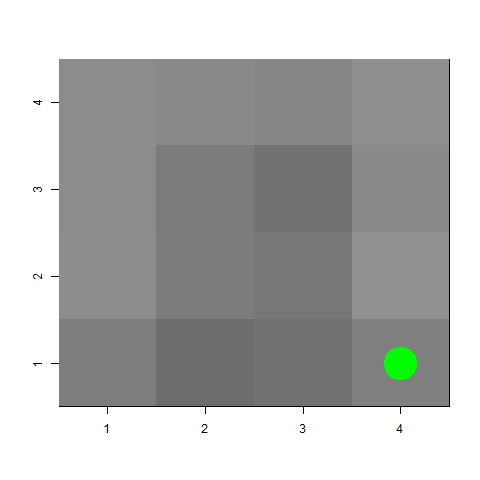}}
\\
\hspace*{-2pt}
\raisebox{0.9cm}{4. }{
\includegraphics[width=2cm, height=2cm]{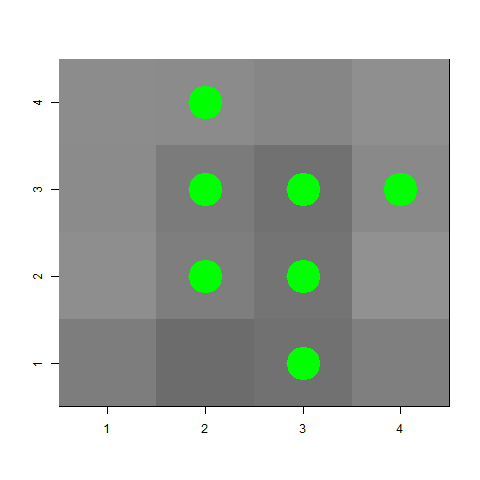}
\includegraphics[width=2cm, height=2cm]{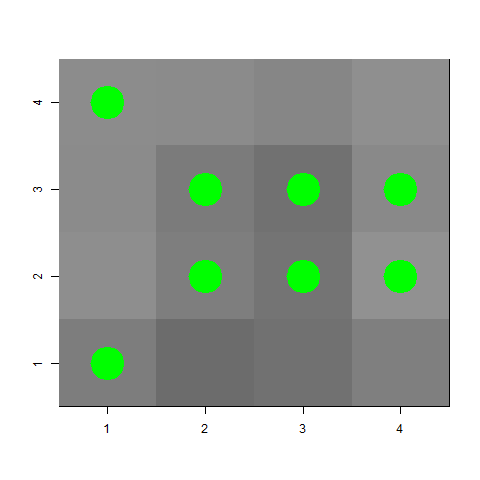}
\includegraphics[width=2cm, height=2cm]{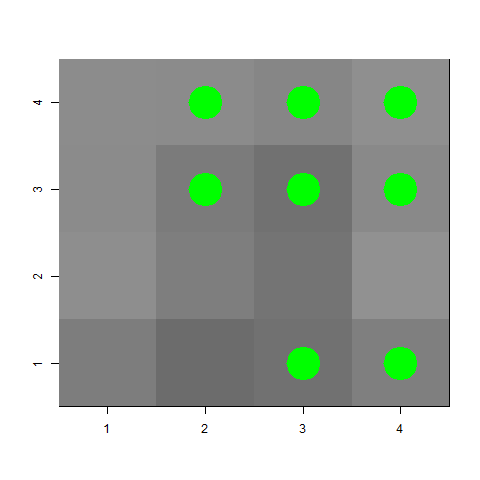}
\includegraphics[width=2cm, height=2cm]{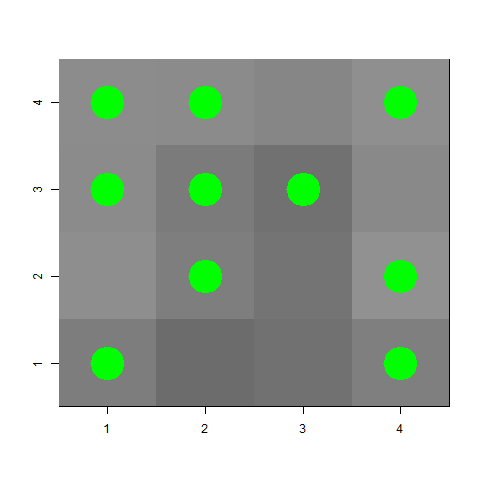}
\includegraphics[width=2cm, height=2cm]{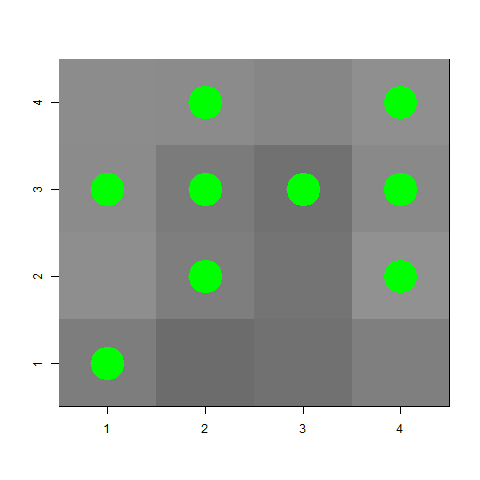}
\includegraphics[width=2cm, height=2cm]{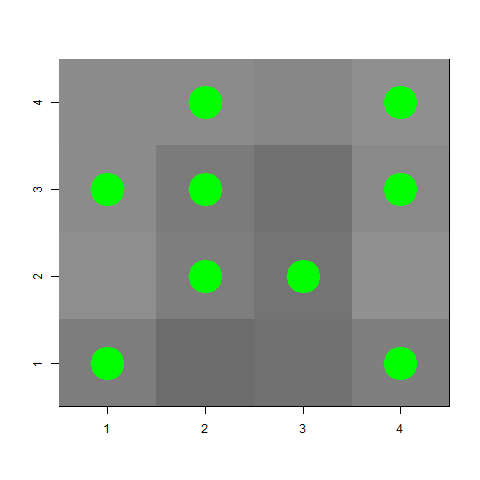}}\\
\vspace{-4pt}\small{(a) a processed boxing video}\\
\hspace{2pt}
\raisebox{0.9cm}{1. }{
\includegraphics[width=2cm, height=2cm]{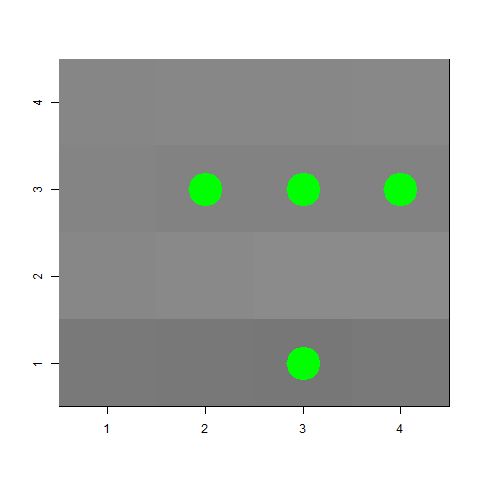}
\includegraphics[width=2cm, height=2cm]{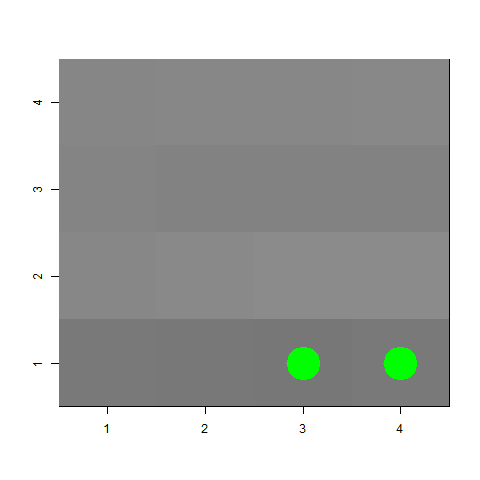}
\includegraphics[width=2cm, height=2cm]{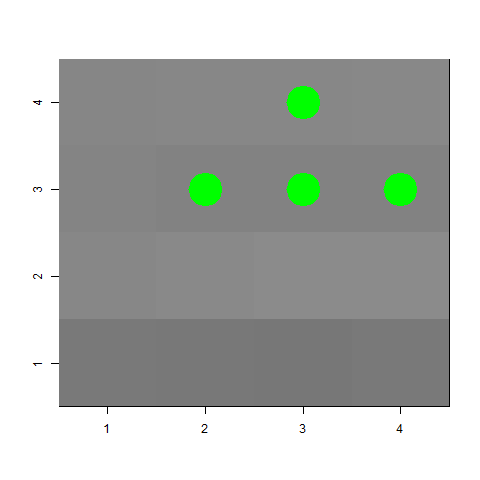}
\includegraphics[width=2cm, height=2cm]{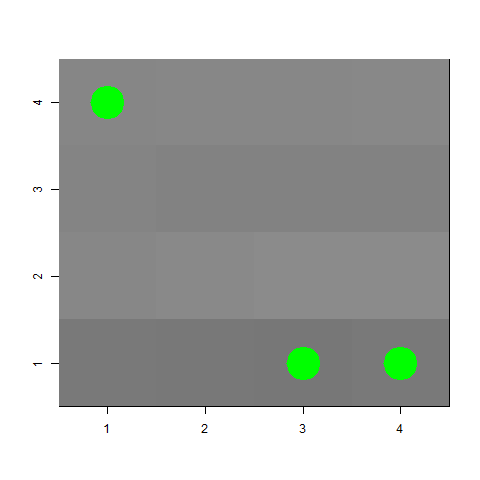}
\includegraphics[width=2cm, height=2cm]{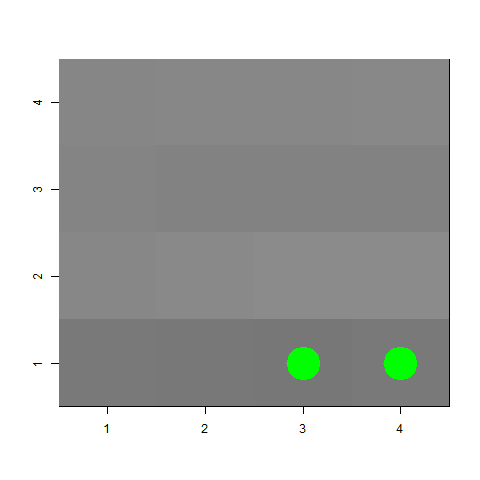}
\includegraphics[width=2cm, height=2cm]{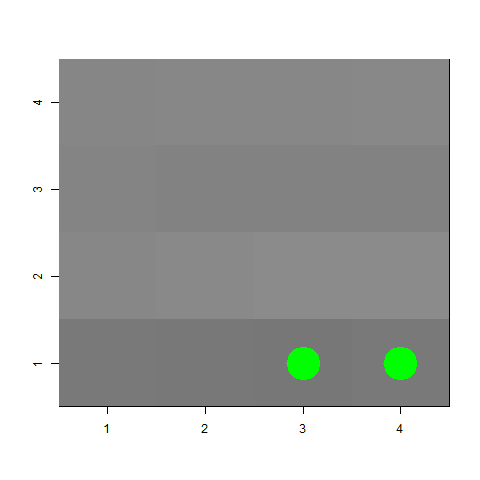}}
\\
\hspace{2pt}
\raisebox{0.9cm}{2. }{
\includegraphics[width=2cm, height=2cm]{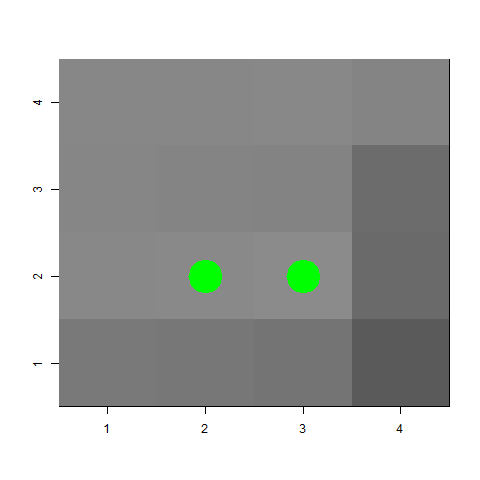}
\includegraphics[width=2cm, height=2cm]{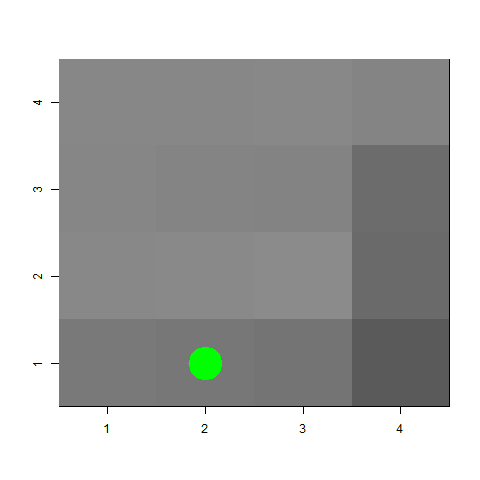}
\includegraphics[width=2cm, height=2cm]{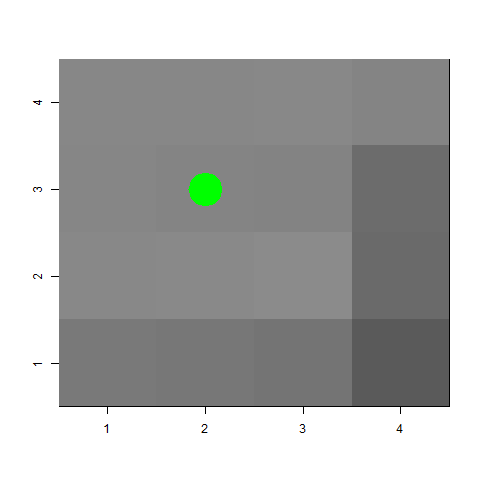}
\includegraphics[width=2cm, height=2cm]{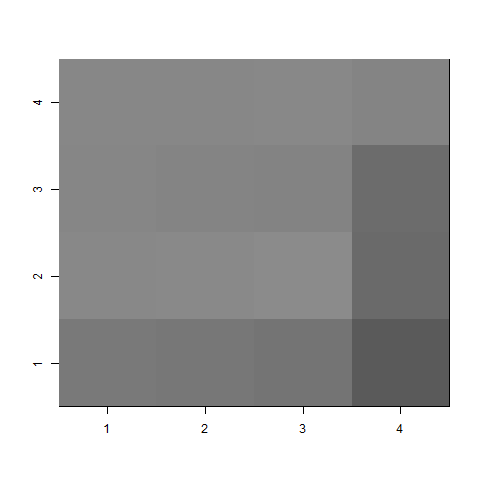}
\includegraphics[width=2cm, height=2cm]{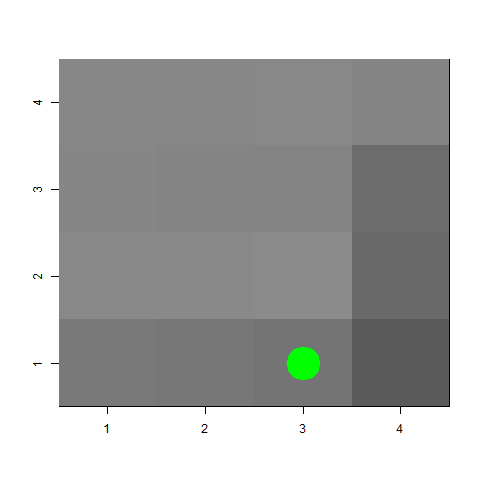}
\includegraphics[width=2cm, height=2cm]{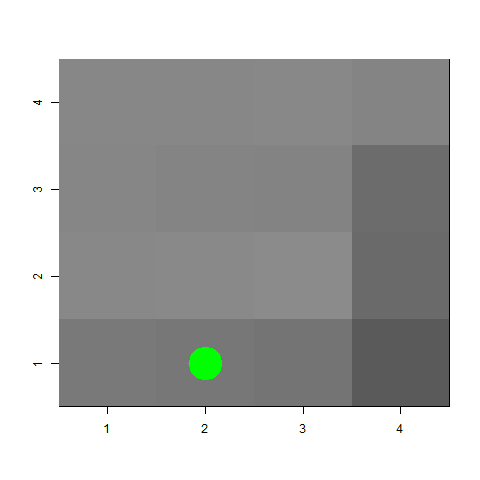}}
\\
\hspace*{2pt}
\raisebox{0.9cm}{3. }{
\includegraphics[width=2cm, height=2cm]{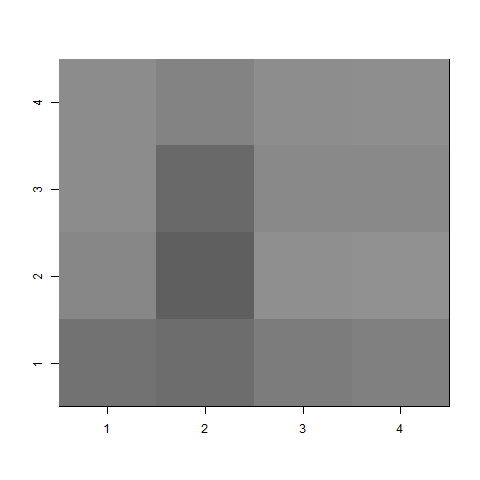}
\includegraphics[width=2cm, height=2cm]{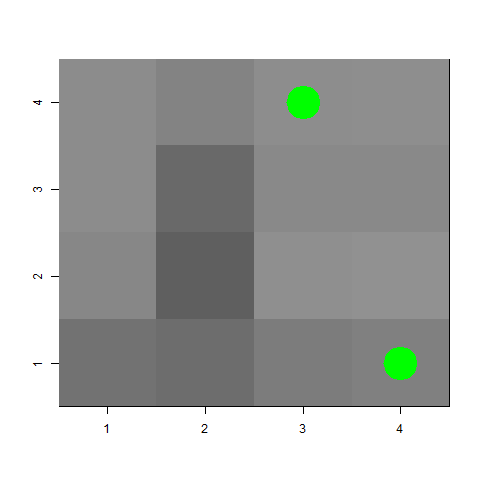}
\includegraphics[width=2cm, height=2cm]{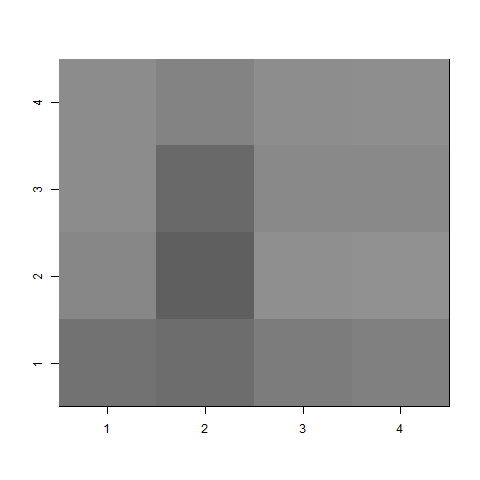}
\includegraphics[width=2cm, height=2cm]{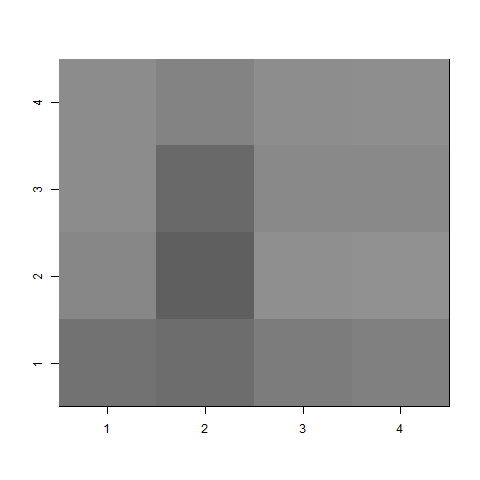}
\includegraphics[width=2cm, height=2cm]{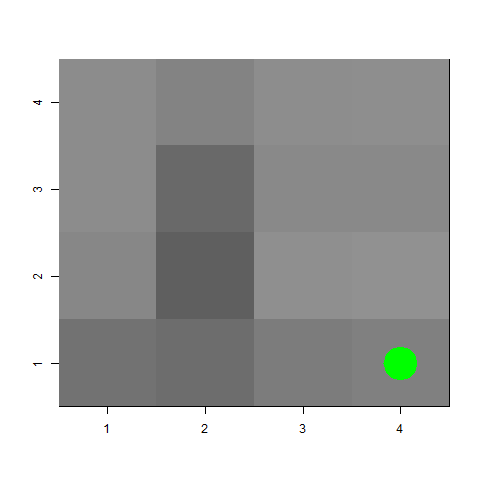}
\includegraphics[width=2cm, height=2cm]{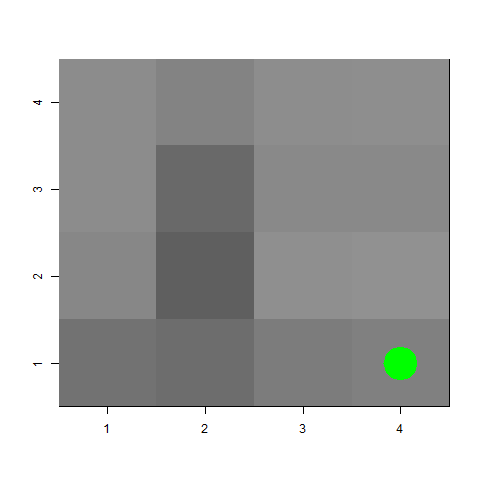}}
\\
\hspace*{-2pt}
\raisebox{0.9cm}{4. }{
\includegraphics[width=2.1cm, height=2.1cm]{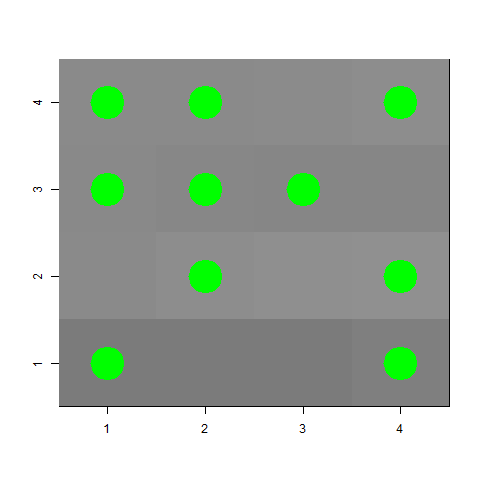}
\includegraphics[width=2.1cm, height=2.1cm]{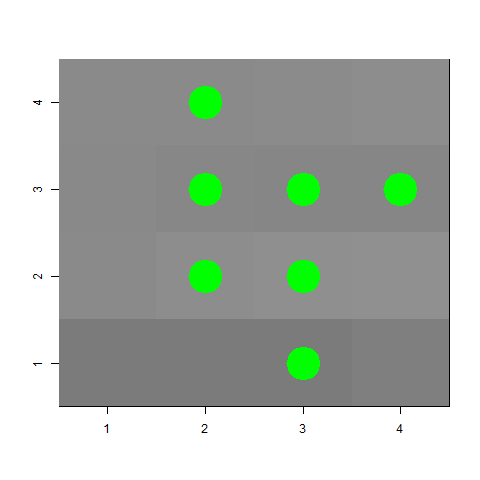}
\includegraphics[width=2.1cm, height=2.1cm]{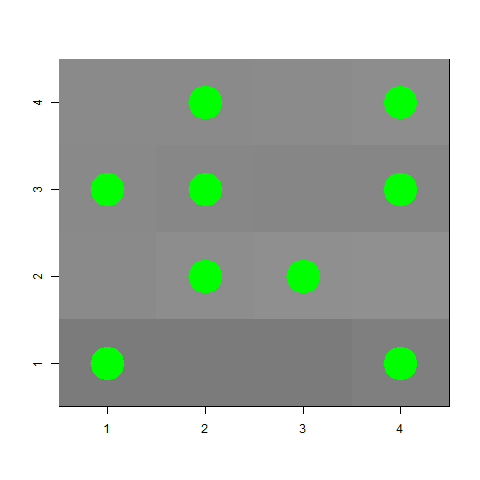}
\includegraphics[width=2.1cm, height=2.1cm]{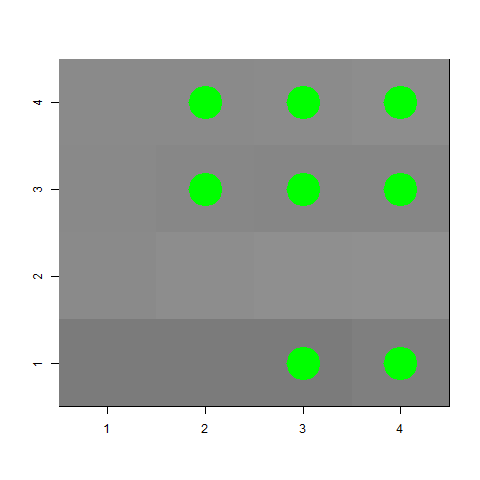}
\includegraphics[width=2.1cm, height=2.1cm]{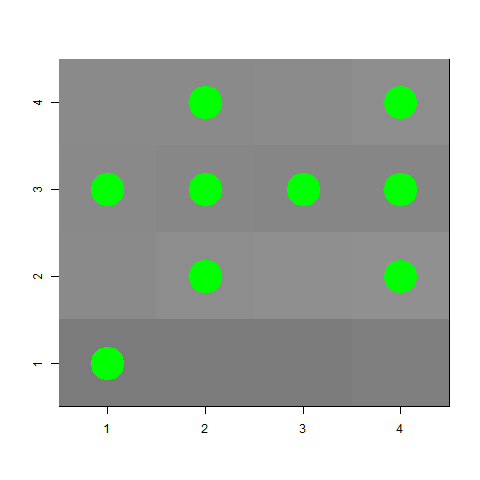}
\includegraphics[width=2.1cm, height=2.1cm]{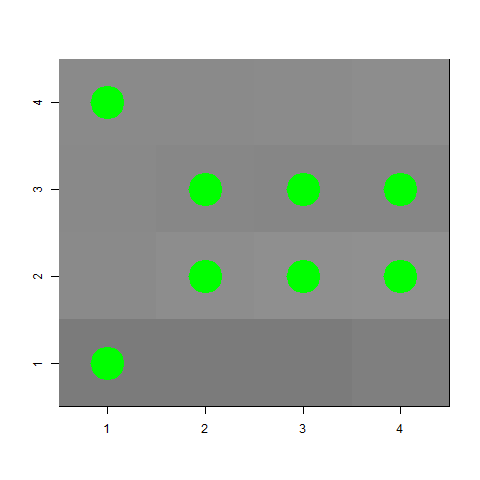}}\\
\vspace{-3pt}
\centering\small{(b) a processed jogging video}\vspace{-3pt}
\caption{Pixels (green dots)  corresponding to the top 20\% elements  in $|\hat{\mathcal{B}}|$ estimated in each methods in two example motion videos. Each row represents a single frame in videos (a) and (b), respectively. The darker vertical ``post'' that appears in many images represents a human subject; it does not move in each frame in video (a), reflecting that the subject stands still, doing hand motion (boxing); it moves in video (b), representing that the subject is jogging into the frame from the right and out of the frame from the left. }
\label{fig:motion}\vspace{-12pt}
\end{figure}

We calculated the misclassification rate in each method on the motion categorization in the testing data and present 
the results in Table \ref{tab:motion}.  We took the top 20\%  elements in the estimated $\hat{\mathcal{B}}$ in absolute value in each methods and plotted the corresponding pixels (green dots) in Figure \ref{fig:motion} for two  videos from a randomly selected subject. 
In term of classification accuracy, \our's performance is in the middle, worse than  $\ell_2$-regularized CP and elastic net (EN)-regularized Tucker, similar to $l_{1}$ regularized GLM, and better than the 3 unregularized models. On the other hand, the image areas relevant for classification identified by \our make the most sense among all the methods. Specifically, first, the selected pixels concentrated on frames 4 and 1 the most, indicating these two frames were the most important frame in terms of distinguishing the two types of motion. This is indeed the case, frames 4 and 1 have either no human or the human is located in the most right or left part of the images in videos with lower body motions. In contract, the human always shows up in videos with upper body motions.  \our is the only method that did not select any pixel from frames 2 and 3, indicating that its ability to focus on the most important frames. second, the middle two column pixels in each frames are also important for distinguishing the two motion types. The human is always in the middle of an image in videos containing upper body motions, and the human's position across images in the sequence in videos containing lower body motions. \our selected a total of 5 pixels  while the other methods picked 6 to 11 pixels  from the middle two pixel columns in all frames, indicating that \our can identify the most important positions in a single frame as well. Taken together, the results from \our were the most interpretable.

\begin{table}[!htb]\vspace{-6pt}
\centering\caption{Motion misclassification rate by different models}\label{tab:motion}\vspace{-6pt}
\resizebox{1\textwidth}{!}{
\begin{tabular}{c@{\hspace{3pt}}c@{\hspace{5pt}}c@{\hspace{6pt}}c@{\hspace{6pt}}c@{\hspace{6pt}}c@{\hspace{3pt}}c}
\hline
$\ell_2$-regularized   & EN$^\dagger$-regularized & $\ell_1$ regularized  &  \our & unregularized   & GLM  & unregularized \\
 CP & Tucker & GLM  on vec($\mathcal{X}$)   &  & Tucker &  on vec($\mathcal{X}$) &  CP\\
\hline
\textbf{1.96\%} & 2.49\%  & 3.20\% &  3.38\%   & 4.98\% & 4.98\% &  5.16\%\\
\hline
\multicolumn{7}{l}{\small{$^\dagger$ EN: elastic net}}\\
\hline
\end{tabular}}\vspace{-6pt}
\end{table}

\section{Discussion}\label{sec:discu}\vspace{-3pt}
We propose \our\hspace{-4pt}, a regularization method for TR through noise augmentation, coupled with low-rank decomposition on the parameter tensor $\mathcal{B}$. We establish theoretically that \our promotes orthogonality between the core tensor and augmented noisy data and achieves exact $\ell_0$ regularization in linear and GLM TR on the core tensor in the Tucker decomposition of $\mathcal{B}$. To our knowledge, our method is the first Tucker decomposition-based regularization method in TR to achieve $\ell_0$ in core tensors. 

\our is implemented through an iterative procedure. Each iteration involves two simple steps -- generating noisy data based on the core tensor estimates from the Tucker decomposition of the $\mathcal{B}$ from the last iteration  and running a regular GLM on noise-augmented data with vectorized predictors. Parameter estimates from \our are robust to the choice of hyperparameter $\lambda$ while $n_e$ requires careful tuning. \our outperforms other decomposition-based regularization methods for TR in both simulated data and real-life data applications and can also identify important predictors though it is not designed for that purpose. 

Due to the cubic time complexity  in the dimensionality $\mathbf{B}$ and  $\mathbf{G}$ in Tucker decomposition and tensor reconstruction in each iteration of the algorithm, we limited the applications of \our to tensor data of relatively low dimensions in the empirical evaluations. We will continue to seek more computationally efficient approaches for Tucker decomposition and tensor re-construction so to apply \our to higher-dimensional tensor datasets, such as medical imaging data, and examine its effectiveness in promoting sparsity in such settings. Since the cubic time complexity is not unique to \our but also arises in other procedures involving  Tucker decomposition and tensor re-construction while the noise augmentation and the GLM step in \our  is  associated only with linear time complexity, one  can be easily code and incorporate the noise augmentation step in software packages and apps that already run tensor regression with Tucker decomposition to help achieve $\ell_0$ regularization for the regression.

We will also explore the asymptotic distributions of the estimated core tensor and the estimated $\mathcal{B}$ via the \our procedure, based on which confidential intervals of the parameters can be constructed. In addition, it would be of interest to quantify the convergence rate of the \our procedure in $n$, the number of parameters in $\mathcal{B}$, and the number of zero elements in the core tensor, and the fluctuations around the final estimates of the estimated parameters upon convergence.

\subsection*{Data and Code}
The data and the code for the \our  procedure, the simulation study, and the real data applications are available at \url{https://github.com/AlvaYan/NA-L0-TR}.

\bibliographystyle{plainnat}
\bibliography{ref}

\begin{appendix}
\section*{Appendix}
\subsection*{A\hspace{12pt}Convergence of \our}\label{sec:appA}
We examine the convergence behavior of \our by presenting the trace plot of the residual deviance (the loss function in GLM) and the parameter estimates over iterations in the \our algorithm in simulated studies. The data were simulated in a similar manner as in Section \ref{sec:simData} except that instead of generating 8 elements which were replicated 8 times to generate $\mathcal{B}$, 32 elements were generated  -- from $\mathcal{N}(0,1)$ in the linear and logistic cases and Unif(0, 0.3) for the Poisson case -- and then replicated twice to construct $\mathcal{B}$, leading to 32 zeros in the core tensor of $\mathcal{B}$. The reason for the change from the simulation settings in Section \ref{sec:simData} is that \our converged very fast there with the large number of zeros (62), making it less ideal to examine the impact of $m$ on the convergence. We decrease the number of zeros to better examine the convergence. The simulation of $\mathbf{X}$ and $Y$ remain the same as in Section \ref{sec:simData}. We set $n_e = 32$ and examined different values of $m$.

\begin{figure}[!htb]
\centering
\hspace{-3pt}\subcaptionbox{\scriptsize{linear}}{
\includegraphics[scale=0.29]{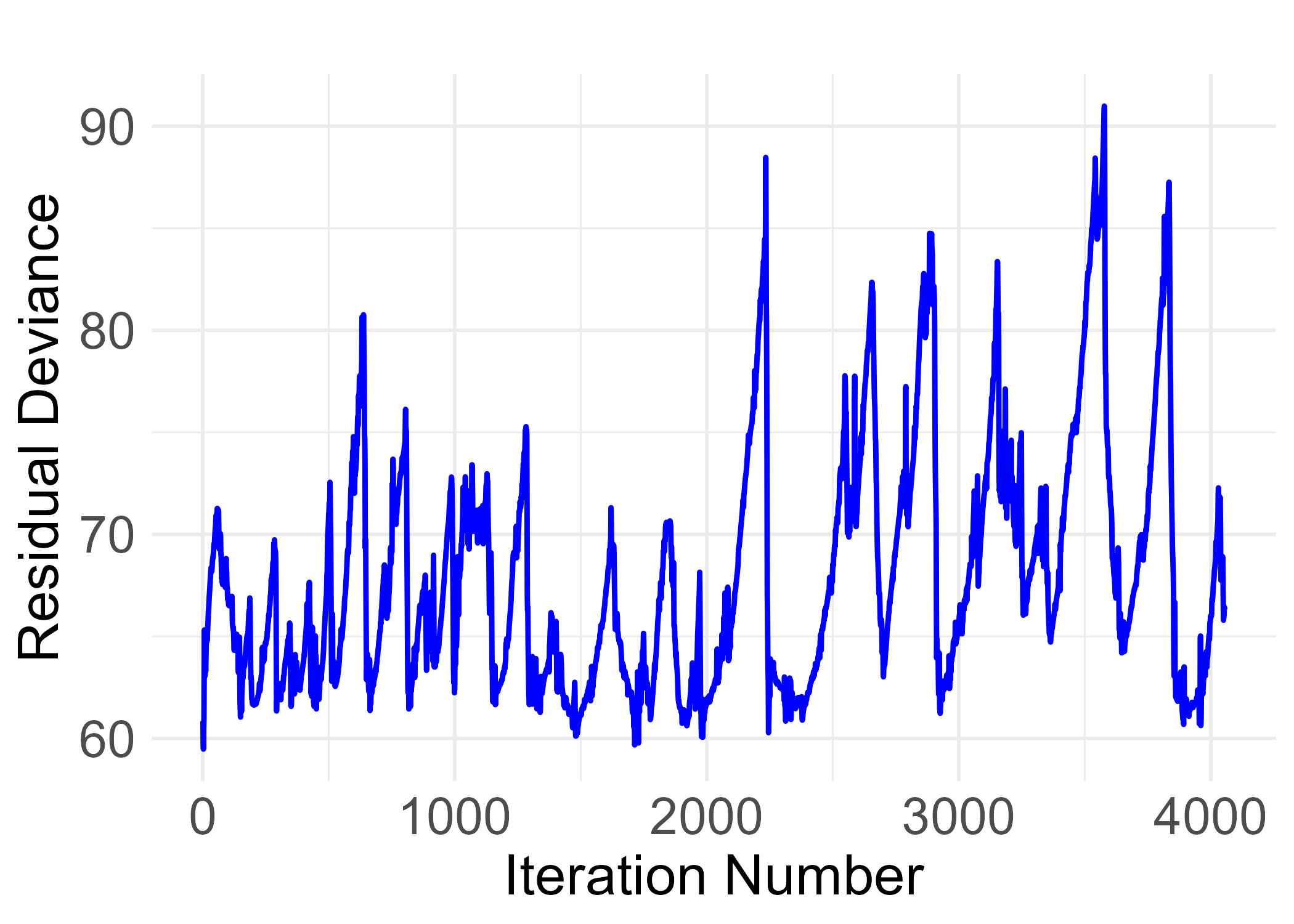}}
\subcaptionbox{\scriptsize{binomial}}{
\includegraphics[scale=0.29]{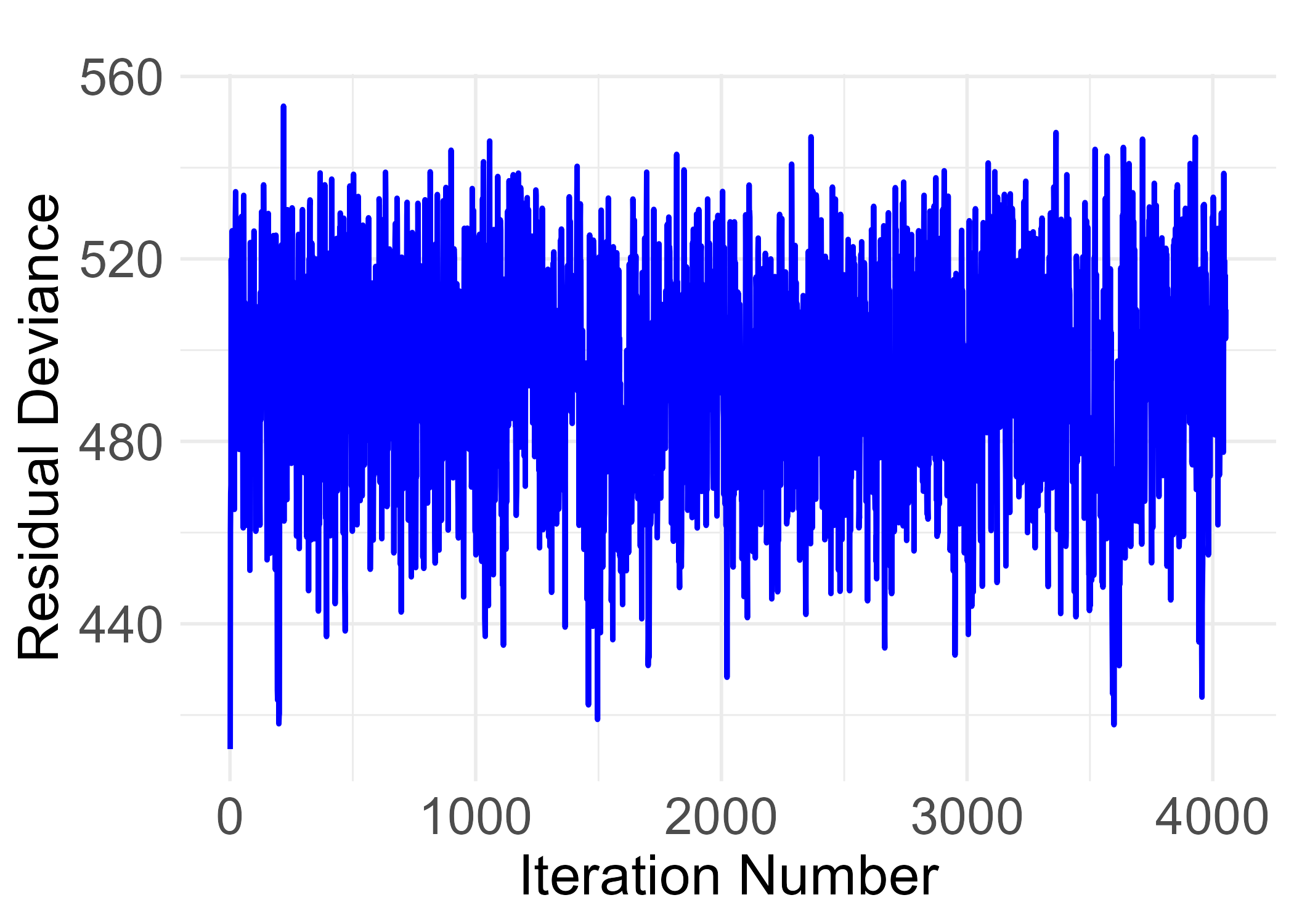}}
\subcaptionbox{\scriptsize{Poisson}}{
\includegraphics[scale=0.29]{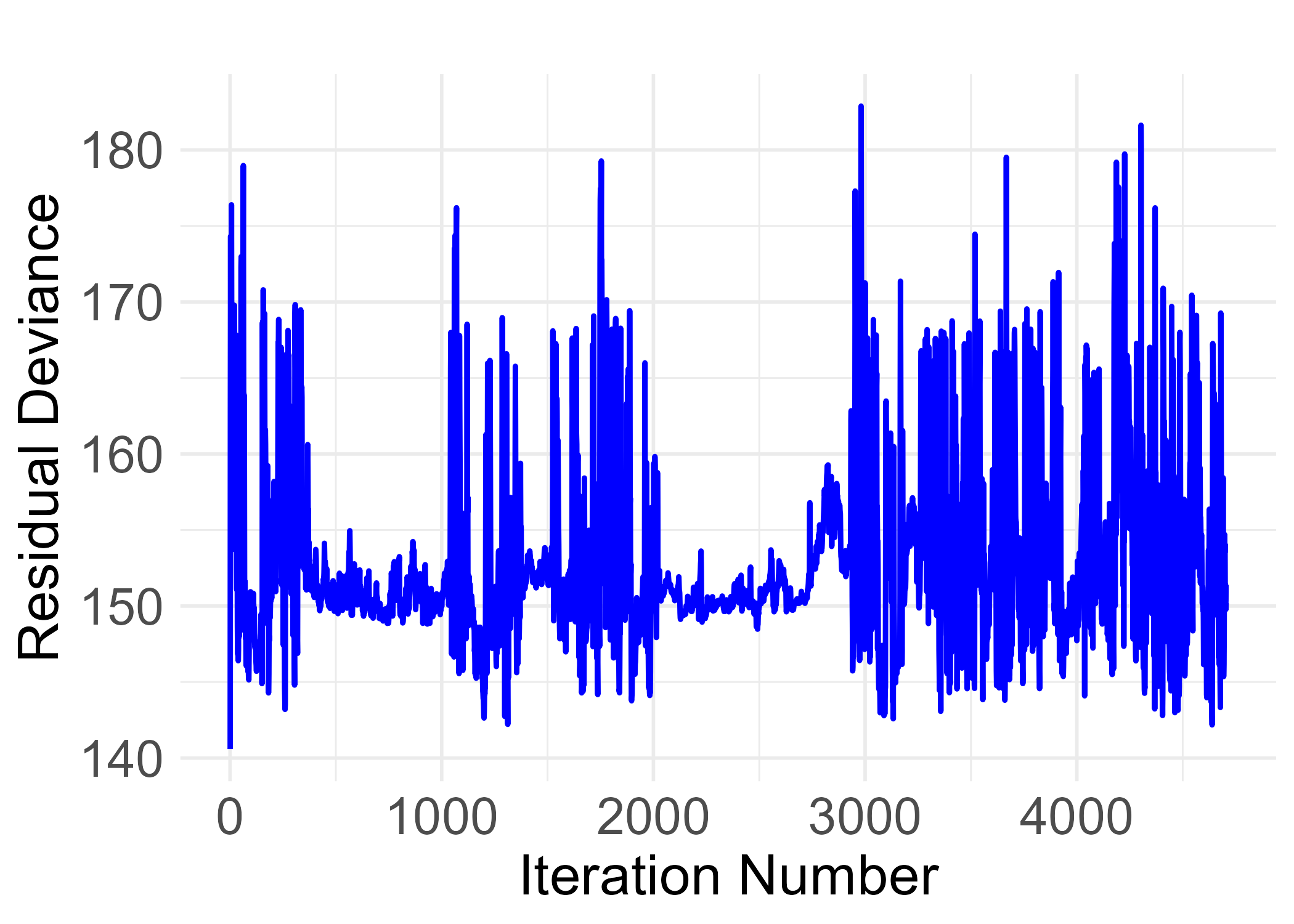}}
\vspace{-3pt}
\caption{Example trace plot of residual deviance in a single repeated dataset in the \our algorithm with $m=1$.}
\label{fig:DVcvgm1}\vspace{-9pt}
\end{figure}

\begin{figure}[!htb]
\centering
\hspace{-3pt}\subcaptionbox{\scriptsize{linear}}{
\includegraphics[scale=0.29]{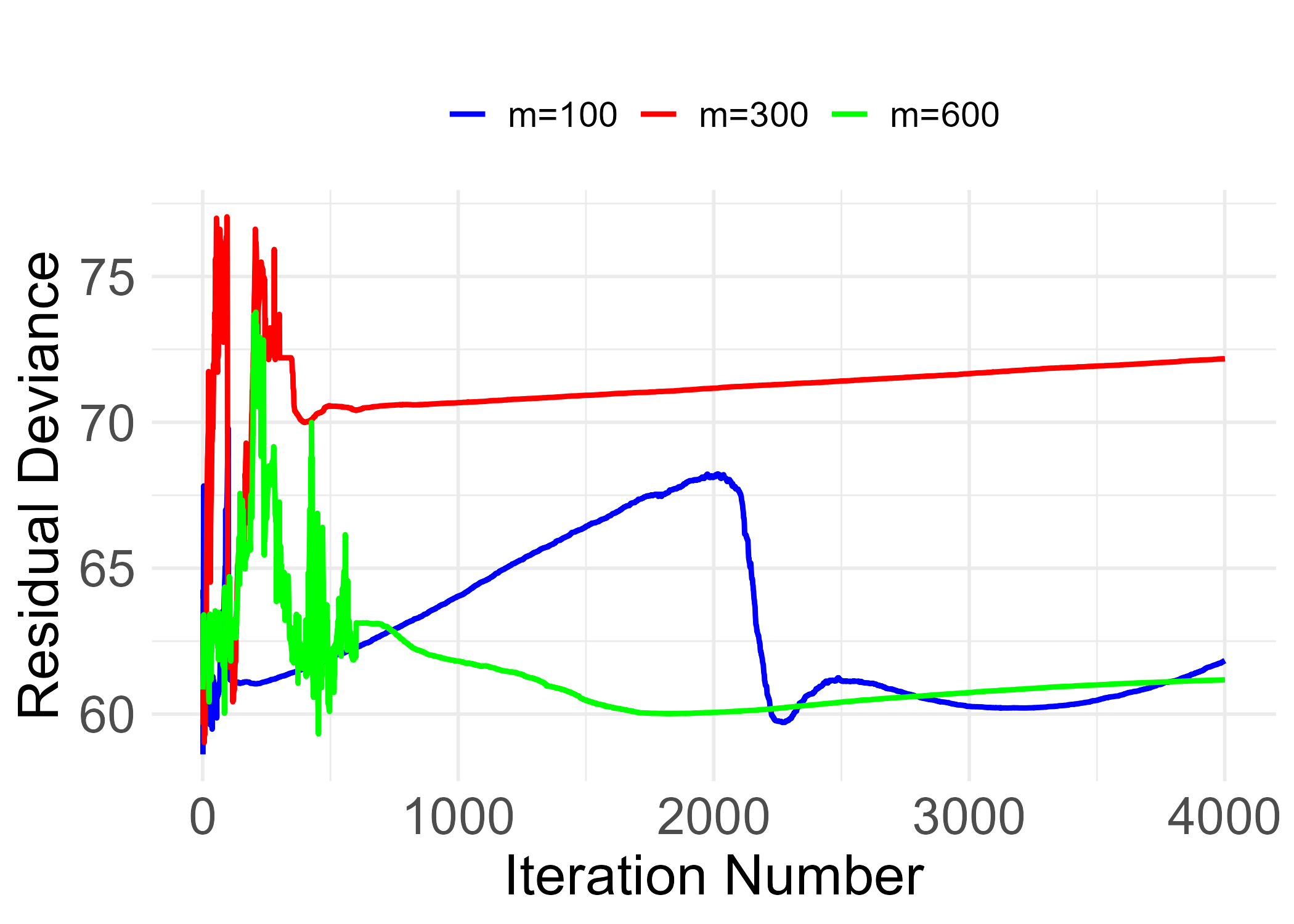}}
\subcaptionbox{\scriptsize{binomial}}{
\includegraphics[scale=0.29]{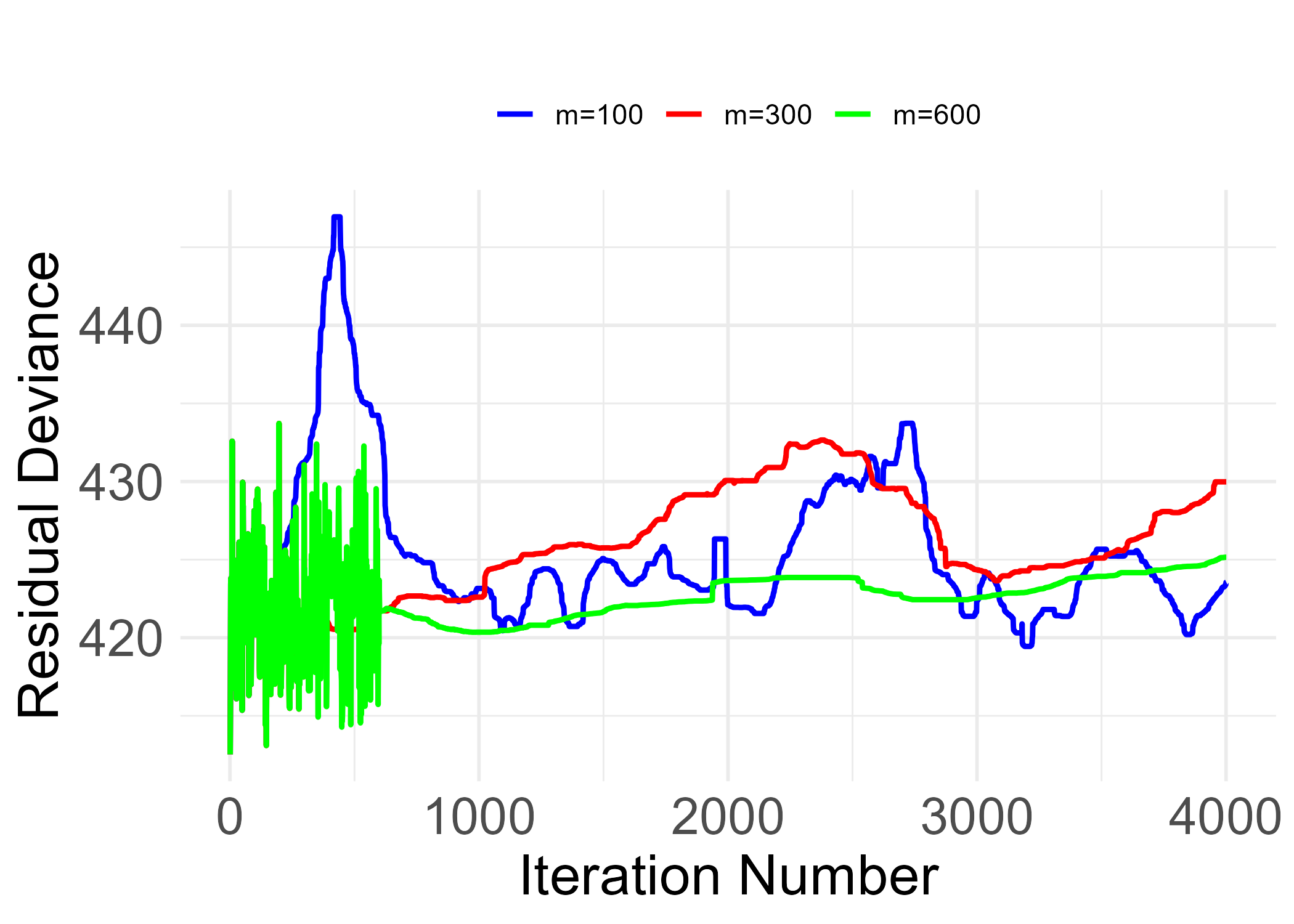}}
\subcaptionbox{\scriptsize{Poisson}}{
\includegraphics[scale=0.29]{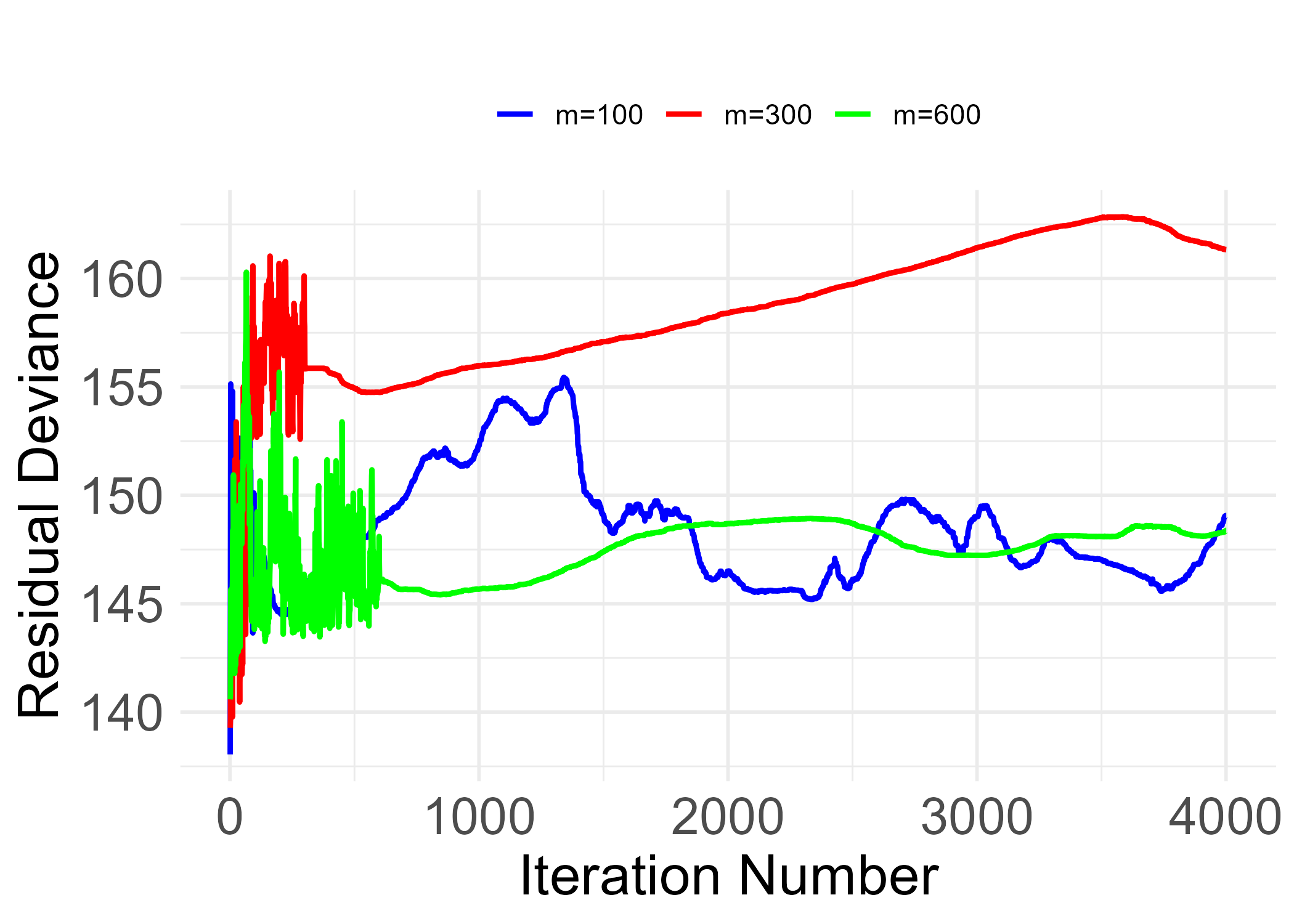}}
\vspace{-3pt}
\caption{Example trace plot of residual deviance in a single repeated dataset 
 in the \our algorithm. The smoothing/averaging  of residual deviance over $m$ iterations does not take place until after the $m$-th iteration has been completed.}
\label{fig:DVcvg}\vspace{-9pt}
\end{figure}

The trace plots of the residual deviance are presented in Figure  \ref{fig:DVcvgm1} for $m=1$  and Figure \ref{fig:DVcvg} for $m=100, 300, 600$.  When there is smoothing over iterations $(m=1)$ (Figure  \ref{fig:DVcvgm1}), the model residual deviation fluctuates significantly -- due to that the noisy data used to augment the original data is a random set generated independently and differ from iteration to iteration. With the data changing from iteration to iteration, so are the parameter estimates based on the changing data, the fluctuation around the residual deviance is expected given that it is a function of both the data and parameter estimates.
When $m>1$ (Figure \ref{fig:DVcvg}), as expected, larger $m$ exhibits less oscillations and more stability in the loss function.  $m=600$ given much more stabilized loss function compared to $m=1$. 

\vspace{-6pt}

\vspace{-6pt}

\subsection*{B\hspace{12pt}\texorpdfstring{Effect of $n_0$ on Estimated Number of Zeros in Core Tensor}}
\label{sec:appB}
In \our, when the regularization parameter $\lambda$ is sufficiently small but still large enough to provide effective regularization,  the number of zeros in the core tensor identified by our method closely approximates the true number of zeros. As $\lambda$ becomes large, the number of zeros estimated in the core tensor eventually equals the size of $n_e$. This insight is particularly useful for selecting appropriate values of $n_e$ and $\lambda$. 

\begin{figure}[!htb]
\centering\vspace{-18pt}
\includegraphics[scale=0.5]{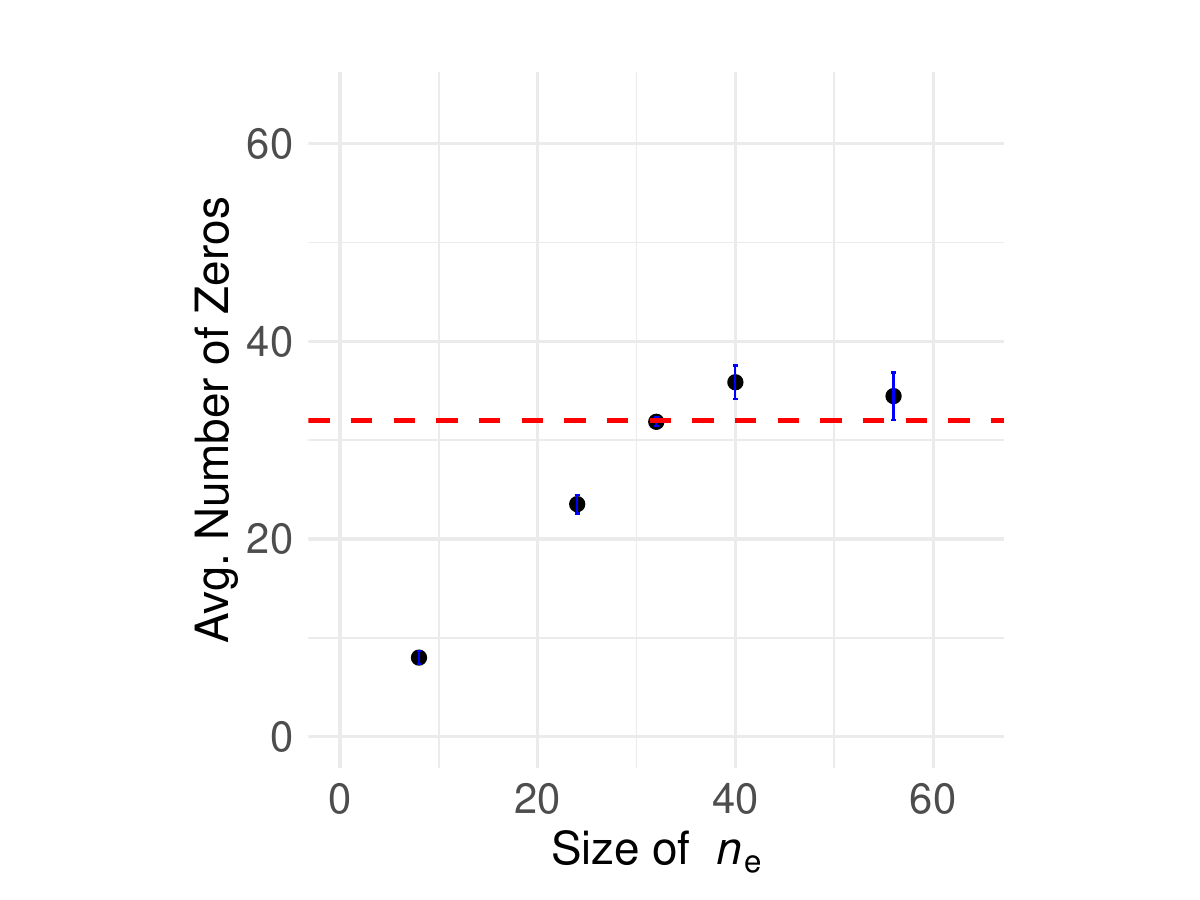}
\vspace{-3pt}
\caption{$n_e$ vs.~average ($\pm$SD) estimated number of zeros in the core tensor via \our in linear TR. The dashed line represents the true number of zeros (32) in the core tensor.}
\label{fig:ne0}
\vspace{-9pt}
\end{figure}

To illustrate this behavior, we conducted an experiment in linear TR in simulated data. We generated $\mathcal{B}$ in the same manner as in Appendix A with $n_0=32$ zeros in the core tensor. In \our\hspace{-3pt}, we set $n_e$ at $8, 24, 32, 40, 56$. For each value of $n_e$, we ran the \our algorithm 100 times with $\lambda=0.01$ and present  the average and standard deviation of the estimated number of zeros in the core tensor over the 100 repeats in Figure \ref{fig:ne0}. When $n_e$ is set at a value smaller than true number of zeros (32), the number of estimated zeros roughly equals to $n_e$. Once $n_e$ exceeds the true number of zeros in the core tensor (32), the estimated number of zeros remains around $n_0=24$ regardless of $n_e$.
\end{appendix}

\end{document}